\RequirePackage{fix-cm}
\documentclass[smallextended]{svjour3}       % onecolumn (second format)
\smartqed  % flush right qed marks, e.g. at end of proof
\usepackage{graphicx}
\usepackage{times}
\usepackage{url}
\usepackage{latexsym}
\usepackage{color}
\usepackage{soul}
\usepackage{array}

\usepackage{multirow}
\usepackage{clrscode}
\usepackage{natbib}
\usepackage{amsmath}

\DeclareMathOperator*{\argmax}{\arg\!\max}

\makeatletter
\def\s@btitle{\relax}
\def\subtitle#1{\gdef\s@btitle{#1}}
\def\@maketitle{%
  \newpage
  \null
  \vskip 2em%
  \begin{center}%
  \let \footnote \thanks
    {\LARGE \@title \par}%
                \if\s@btitle\relax
                \else\typeout{[subtitle]}%
                        \vskip .5pc
                        \begin{large}%
                                \textsl{\s@btitle}%
                                \par
                        \end{large}%
                \fi
    \vskip 1.5em%
    {\large
      \lineskip .5em%
      \begin{tabular}[t]{c}%
        \@author
      \end{tabular}\par}%
    \vskip 1em%
    {\large \@date}%
  \end{center}%
  \par
  \vskip 1.5em}
\makeatother 

\begin{document}

\title{Harmonic Grammar, Optimality Theory, and Syntax Learnability:}
\subtitle{An Empirical Exploration of Czech Word Order}
\author{Ann Irvine \and Mark Dredze}
\date{}
\institute{Ann Irvine and Mark Dredze \at
	     Department of Computer Science, Johns Hopkins University \\
              3400 N. Charles St. \\
              Baltimore, MD 21218 \\
              Tel.: 1-540-460-3663\\
              \email{annirvine@gmail.com, mdredze@cs.jhu.edu}           %  \\
%             \emph{Present address:} of F. Author  %  if needed
%           \and
   %        Mark Dredze \at
%	     Johns Hopkins University \\
%              3400 N. Charles St. \\
%              Baltimore, MD 21218 \\
%              \email{mdredze@cs.jhu.edu}           %  \\
}

\maketitle

\begin{abstract}This work presents a systematic theoretical and empirical comparison of the major algorithms that have been proposed for learning Harmonic and Optimality Theory grammars (HG and OT, respectively). By comparing learning algorithms, we are also able to compare the closely related OT and HG frameworks themselves. Experimental results show that the additional expressivity of the HG framework over OT affords performance gains in the task of predicting the surface word order of Czech sentences. We compare the perceptron with the classic Gradual Learning Algorithm (GLA), which learns OT grammars, as well as the popular Maximum Entropy model. In addition to showing that the perceptron is theoretically appealing, our work shows that the performance of the HG model it learns approaches that of the upper bound in prediction accuracy on a held out test set  and that it is capable of accurately modeling observed variation.
\end{abstract}

\vspace{10 mm}


\clearpage

\section{Introduction}

A complete theory of grammar must explain not only the structure of linguistic representations and the principles that govern them but also how those grammars are learned. Most of the work in Optimality Theory (OT)  \citep{Prince1993,prince2004} has focused on discovering the constraint rankings that speakers use to produce and understand a given language, and how such rankings vary across languages. However, there have been several threads of research on modeling how speakers learn the constraint rankings in the first place. The earliest of the learnability research focused on OT grammars \citep{tesarsmolensky1993, tesarthesis, Tesar1998, Tesar2000, Boersma1997, Boersma2001}. However, recently there has been a resurgence of interest in the Harmonic Grammar (HG) framework \citep{Legendre1990, legendre1990a, legendre1990b, harmonicmindchaptera, harmonicmindchapterb, coetzee08a, pater2009, jesney2009, potts2010}, which is closely related to OT. In this work, we theoretically and empirically compare learning algorithms for each and, by doing so, are able to compare the frameworks themselves.

As noted in recent work, including \cite{Pater08}, \cite{boersmaPater08}, and \cite{Giorgio10}, the perceptron learning algorithm is well-established in the Machine Learning field and is a natural choice for modeling human grammar acquisition. The algorithm learns from one observation at a time, and it is capable of learning from a noisy corpus of observed natural language. In this work, we make a case for the perceptron, which we use to learn a model that specifies a set of constraint weights relevant to one syntax phenomenon, Czech word order. We extract training data (sentences annotated with grammatical and information structure and their surface word orders) from the Prague Dependency Treebank \citep{pdt} and use basic alignment (edge-most) constraints on grammatical and information structure to predict the surface order of the subject, verb, and object. The perceptron algorithm learns how to numerically weight a set of constraints -- a Harmonic Grammar. 

Ordering HG constraints by the magnitude of their weights may specify a hierarchical constraint ranking, an OT Grammar, and doing so is the essence of the classic Gradual Learning Algorithm (GLA) \citep{Boersma1997}. Recent work \citep{magri2011} has pointed out that any HG learning algorithm can be adapted to learn an OT grammar and, thus, OT learnability is no more computationally complex than HG learnability. However, HG is a more expressive grammar framework. That is, all OT grammars can be expressed as HG grammars but not vice-versa. In this work, we automatically learn both types of grammars and show that the additional expressiveness of Harmonic Grammars results in gains in surface form prediction accuracies.
%In this work, I use a held out set of empirical data to quantitatively evaluate how well a perceptron can learn both types of grammars and, furthermore, how well each is able to predict production variation. 
%As pointed out by \cite{Giorgio10}, the perceptron can learn both types of grammars, and 
%In addition to the perceptron and GLA, we explore the MaxEnt learning
%We describe and compare the GLA and Perceptron learning algorithms and OT and HG frameworks in detail, and 
That is, we use a held out set of empirical data to quantitatively evaluate each and find that allowing for so-called \emph{ganging-up-effects}, the more expressive Harmonic Grammar models Czech Word Order more accurately than the OT grammar learned by the GLA. Furthermore, the performance of the HG grammar learned by the perceptron approaches that of the upper bound, which is defined in terms of the information available to any learner.

We compare the perceptron and GLA and the prediction accuracies of the grammars they learn with a Maximum Entropy model, another popular model of HG/OT learnability \citep{Jaeger2003, Goldwater2003,HayesWilson08}. Crucially, we show that, relative to the other two algorithms and our baseline strategies, the perceptron-learned grammars capture {\it variation} in production well. 

All previous research on OT/HG learnability has focused on the phonological component of a grammar. To date there have been no thorough studies that explore the learnability of the syntax component of an OT grammar.
%\footnote{This is likely due to the fact that syntactic constraint sets and violations often depend on complex tree structures and derivations, which are unavailable in most corpora, and which are difficult to automatically generate and analyze. This is in contrast to data relevant to OT phonology grammars, where single strings can usually be straightforwardly mapped to constraint violations.} 
In this work we contribute to the ongoing research comparing HG and OT and learning algorithms for each from the perspective of learning the syntax grammar component. In particular, our work:

\begin{itemize}
\item{Explores the learnability of OT/HG syntax.}
\item{Relies on observed data to empirically learn, evaluate, and directly compare grammars and learning algorithms.}
%\item{An online learning method, which updates the syntax grammar by considering a single sentence at a time}
\item{Uses the online perceptron algorithm to learn a Harmonic Grammar (HG).}
\item{Demonstrates that {\it ganging-up-effects} provide HG an empirical advantage over OT in terms of modeling syntax.}
\item{Directly compares the perceptron, the GLA, and a Maximum Entropy (MaxEnt) learner in terms of:}
\begin{itemize}
\item{Their theoretical applicability to the learnability problem,}
\item{How accurately each learned model predicts the word order corresponding to a particular input,}
\item{How accurately each models word order variation observed over a large corpus.}
\end{itemize}
\end{itemize}

Our work shows that the perceptron is both theoretically and empirically attractive for modeling learnability and that Harmonic Grammars provide necessary additional power over OT grammars. Furthermore, we show that it is possible to empirically explore the learnability of the syntax component of HG/OT grammars.

\section{Background}\label{background}

\subsection{Optimality Theory and Harmonic Grammars}\label{sec:otandhg}
We assume that our readers are familiar with the Optimality Theory (OT) framework proposed in \cite{Prince1993} and Harmonic Grammar (HG) proposed in \cite{Legendre1990}, and we only briefly review each here. In OT, candidate surface forms are evaluated based on a hierarchical constraint ranking. Constraints are indicators on some element of a hypothesis surface form and, in some cases, also the input. In comparing a pair of candidates, the optimal candidate makes fewer violations of the highest-ranked constraints which distinguish the two. 
Importantly, constraints are violable. That is, the optimal surface form may violate some constraints. In iterating down through the hierarchy, once a candidate makes a violation that at least one other remaining candidate does not, it is immediately eliminated. If all remaining candidates violate a constraint, the process continues.
The optimal surface form is the {\it most harmonic}.

In OT, harmony is defined by a total ordering over all candidate surface forms. In contrast, in HG, harmony is defined by a score associated with each candidate.
In HG, a candidate's harmony is determined by the sum of weights associated with each constraint that it violates. Low ranked constraints contribute to the harmony of candidates and, thus, may impact the choice of optimal surface form, including in cases where using strictly dominated constraints would have rendered them inconsequential. In other words, lower ranked constraints have the potential to {\it gang-up} on higher ranked constraints. 
A harmonic grammar which uses numeric weights can always express a hierarchical OT grammar. Weights would just need to be different enough that lower weighted constraints could never overpower higher ranked constraints, even cumulatively. The power of two series is one example of a series of weights that have this property: 1, 2, 4, 8, 16, 32, 64, 128, etc.

Although OT was proposed after HG, recently HG has regained research momentum in the community, e.g. \cite{PaterBhattPotts07, coetzee08a, boersmaPater08, pater2009, potts2010}. However, it remains to be seen empirically if it offers advantages over OT to justify its additional complexity. \cite{jesney2009} provide some evidence that ganging-up effects can be empirically observed. We further this claim and give a detailed discussion in Section \ref{sec:gang}.

\subsection{Learnability}

As pointed out in \cite{Johnson2009}, there are two, potentially separate, grammar learning problems. The first challenge is to define the structure of the learned grammar itself. %This work is influenced by and often combined with more general attempts to model speaker (one who has already learned the grammar) production. 
This learning problem is also referred to as {\it learnability}. The second challenge is to explain the particular algorithm or process involved in learning. 
Much of this work is done under the name {\it language acquisition}. 
To date, linguists have addressed both facets of learning, usually focusing on one or the other. 

%Variation in general: Bresnan07, Doyle08

The research literature on the learnability of an OT model of speakers' knowledge of grammar and their production patterns, including variation, is extensive and includes the influential papers by \cite{Jaeger2003} and \cite{Goldwater2003}. In contrast to the plethora of work on learning phonological grammars given both underlying and surface word forms, \cite{Gaja} learns phonological production patterns from surface forms alone. There is also some work in modeling, more specifically, a {\it learner's} production patterns from the perspective of stages of language acquisition \citep{Wilson06,Legendre04}.  The majority of work in this line of research models grammar using Maximum Entropy models, which are naturally capable of producing expected probability distributions over possible surface forms. %not only the feature weights corresponding to constraints, but also the underlying forms that make up the lexicon, observing only surface forms of several phonological phenomena.

Somewhat separate are the efforts to model the particular learning algorithm that allows humans to acquire the grammar of a language. Since humans learn incrementally, from individual words and sentences in sequence, this line of research has focused on computational learning models that also learn from a sequence of individual words and sentences. The machine learning community refers to these algorithms that only look at one training example at a time as {\it online}. Algorithms that consider all of the training data available to them at once are, in contrast, called {\it batch} algorithms.%\footnote{Most standard algorithms for learning a Maximum Entropy model are done in batch, not online.}

\cite{Tesar1998, Tesar2000} present the first set of algorithmic discussions. They propose an algorithm called Constraint Demotion (CD) and provide manual simulations of the process of learning a variety of phonological constraint rankings. 
%They also describe the amount and type of data that would be needed to learn a full ranking. Throughout this preliminary work, they assume that the learner has access to the full structural descriptions underlying observed surface forms. 
\cite{eisner00} discusses the computational complexity of the CD algorithm and extensions of it. \cite{Boersma1997} and \cite{Boersma2001} present the Gradual Learning Algorithm (GLA), which learns an Optimality Theory grammar and is an alternative to the CD algorithm. In contrast to CD, the GLA may learn a {\it Stochastic Optimality Theory} (SOT) grammar, which can account for noisy input data and was proposed as way to learn production variation. 
%Boersma and Hayes, Pater, Tesar and Smolensky on learning algorithm itself. The first group addresses language variation. The second addresses specific learning algorithms, and confirms that what is learned is theoretically sound and, at a very basic level, what we would expect. 
\cite{Pater08} points out a flaw in the GLA algorithm and gives a high level discussion of the perceptron learning algorithm as a possible alternative. \cite{boersmaPater08} extend this idea and propose a Harmonic Grammar version of the GLA algorithm (HG-GLA), which is similar to the perceptron algorithm. The HG-GLA work provides a detailed theoretical exploration of the learning algorithm, including some learning computer simulations. \cite{coetzee08a,coetzee08b} use the HG-GLA learner to model phonological variation. We explore each of these algorithms in more detail in Section \ref{sec:learningcomp}.

\section{Motivation}

This work brings together the two threads of work on grammar learning described above. It follows the major intuition of the work done on the second, algorithmic challenge: the algorithms that model human learning should learn online, from one sentence at a time. It follows directly from recent theoretical discussions of such algorithms \citep{Pater08, Giorgio10, boersmaPater08} as well as a plethora of work on the properties of the perceptron algorithm \citep{rosenblatt1958}. Additionally, like much of the previous work, it empirically evaluates the models learned by the algorithms, which should accurately predict observed surface forms, including patterns of variation over possible surface forms. Sections \ref{sec:learning} and \ref{sec:results}, below, give systematic theoretical and empirical comparisons of standard algorithms explored in the OT and HG literature including Constraint Demotion (CD), the Gradual Learning Algorithm (GLA), the perceptron, %\footnote{The perceptron is somewhat similar to the HG-GLA but is well-established in many fields and its theoretical properties are well-known.}, 
and the Maximum Entropy (MaxEnt) %\footnote{Algorithms which are not online will be referred to as batch algorithms. These algorithms consider all training data at once rather than example by example, or, in the case of this work, sentence by sentence.} 
model. Throughout this work, we focus on learning from an observed, noisy corpus, and model a challenging grammatical phenomenon, Czech word order. Our major claim is that the perceptron is a good learner for the following reasons:
\begin{itemize}
\item{It is an online (learns from one sentence at a time) learner.}
\item{It is a simple learning algorithm, so it does not require complex models of broader cognition.}
\item{It is well-studied and its properties are well-understood.}
\item{It is accurate because it can allow for ganging-up effects.}
\item{The model it learns does a good job of predicting observed variation.}
\end{itemize}
Along the way, we show empirically that the additional complexity of Harmonic Grammars over Optimality Theory grammars yield higher prediction accuracies.

\section{Czech Word Order}
\subsection{Phenomenon}
Czech is a language with a relatively free word order \citep{naughton2012czech}. That is, it is common to see all orders of subject, verb, and object in grammatical sentences. However, the basic order is SVO, and the surface word order is strongly influenced by the information structure, or discourse, notions of topic and focus. Topic is old information with respect to the current discourse, and it is usually aligned left in the surface order. Focus, on the other hand, is new information and is usually aligned right in the surface order \citep{Hajicova}. In this work, we automatically learn the ranking of constraints that dictate the Czech word order phenomenon.

The relationship between syntax, phonology, and discourse is complex and the subject of much prior work in linguistics, including \cite{lambrecht} and \cite{kiss}. Using Russian as an example language, \cite{king} discusses the interaction between these levels of structural encoding and, in particular, the implications for the scope of information structure elements. Like Russian, surfacing Czech word orders demonstrate the interaction between information structure and syntax structure. 

In this work, we assume that information structure elements play a role in determining the surface word order of Czech sentences. %and make no further theoretical claims about the interaction between information structure and syntax. 
We take one set of annotations, described below, as given, and do not explore the degree to which they are complete and correct. It is important to keep in mind, however, that the learning frameworks that we describe could incorporate any additional or different sets of similar annotations.

\subsection{Data}\label{sec:data}
Our dataset comes from the Prague Dependency Treebank (PDT) \citep{pdt}. The PDT provides a very rich set of annotations, both automatically and manually created. It annotates dependency structure as well as part of speech, morphology, semantic relations, information structure, and more. 

\begin{table}
\begin{center}
\label{table:orderfreqs}
\begin{tabular}{|c|c|}
\hline
Word & Percent of \\
Order & Corpus Sentences \\
\hline
SVO & 50.0\% \\
OVS & 18.3\% \\
VSO & 10.9\% \\
SOV & 10.1\% \\
VOS & 8.3\% \\
OSV & 2.5\% \\
\hline
\end{tabular}
\caption{Basic word order frequencies in the set of 2955 PDT sentences.}
\end{center}
\end{table}

We selected a subset of the sentences in the PDT from which to learn and evaluate a grammar of Czech Word Order. We limited the set to those sentences that are declarative; have only a single, transitive predicate; and which do not contain commas (which often indicate dependent clauses, relative clauses, or information structure functions). Additionally, we trim prepositional phrases from these sentences, which allows us to focus on the subject, verb, and object elements, and their order. The sentences are simple and, thus, simple to learn from. Table 1 shows the frequency of different word order patterns found in a set of 2955 simple, fully annotated, transitive sentences in the PDT. %As discussed in the preceding section, Czech has a relatively free word order, and these frequencies show that. %\footnote{One contribution of this work is that I have written a script to process all four annotation layers, given in separate files, provided by the PDT. The same script filters the sentences based on the criteria described. Finally, it automatically determines the constraint violations for both positive and implicitly negative candidates. This script can be made available upon request.}.

\begin{figure}[h]
\vskip 0.0in
\begin{center}
\includegraphics[width=0.8 \linewidth]{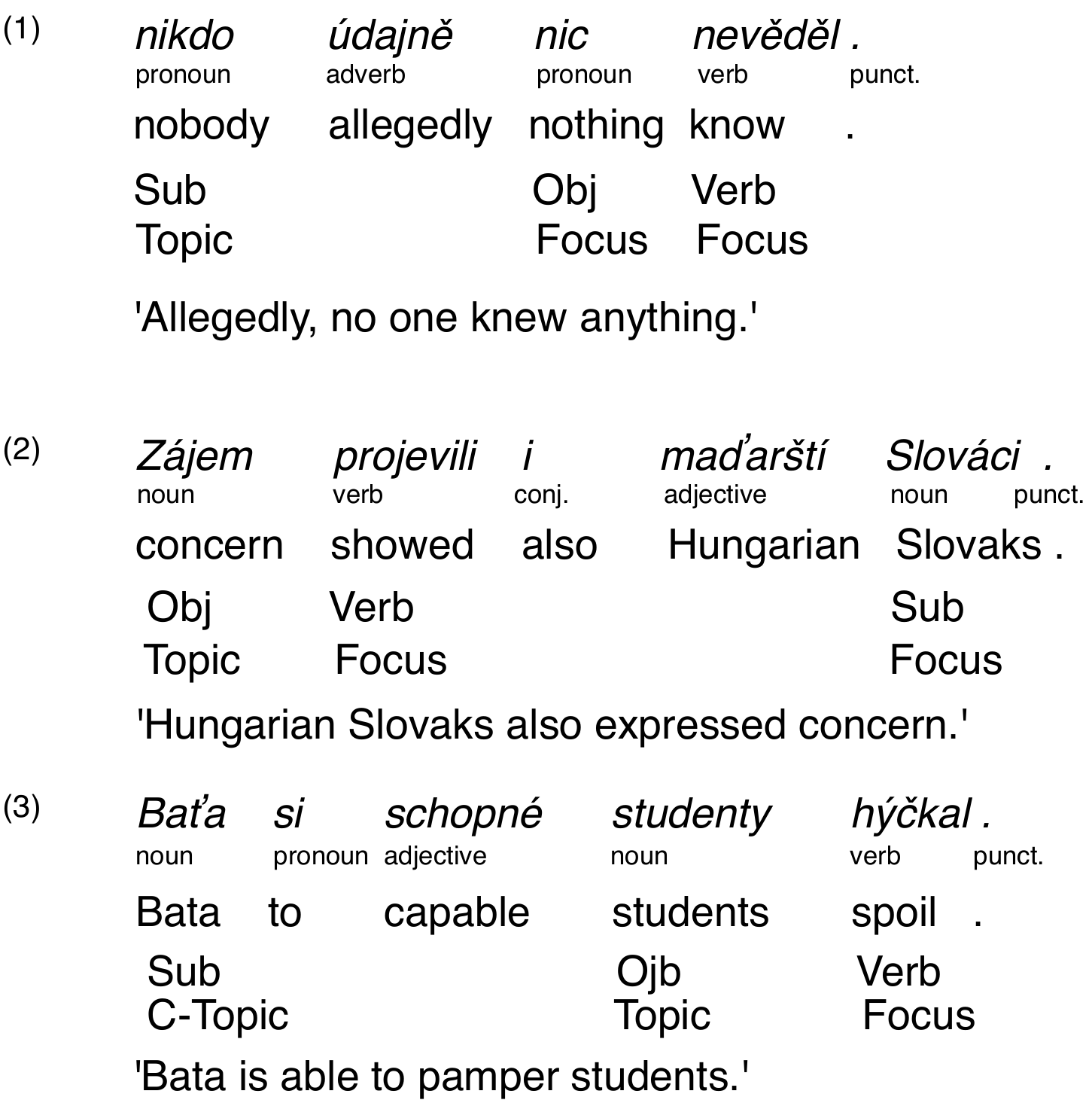}
\caption[]{Sample sentences from the PDT training set, along with their information and grammatical structure annotations. POS tags are given by the PDT. The first and last sentences have SOV order the second has OVS order. \footnotemark }
\label{fig:pdtexamples} 
\end{center}
\vskip -0.2in
\end{figure}

Each word in each sentence in the PDT is annotated with one of the following \citep{Hajicova1998, buranova00}:
\begin{itemize}
\item{t=non-contrastively contextually bound [topic, or old information]}
\item{c=contrastive contextually bound expression [contrastive topic, or contrastive old information]}
\item{f=contextually non-bound expression [focus, or new information]}
\end{itemize}
f, or focus, is the default annotation. Along with grammatical function (S, O, V), these annotations on unordered words form the input to the grammar. Three example sentences are given in Figure \ref{fig:pdtexamples}.

\subsection{Constraint Set}
\label{sec:constraint}
\cite{Choi99} gives a full discussion of word order variation from an Optimality Theory perspective. That work explores German and Korean and suggests OT constraints that place elements with certain salient information structure properties (e.g. newness, prominence) in particular positions in the sentence (e.g. preverbal, sentence initial). We refer readers to Choi's detailed review of the interaction between constituents and their information and discourse structure properties. The algorithms that we discuss in this work could learn rankings of any proposed, violable OT constraints.

%figure footnote text
\footnotetext{Note that the PDT annotations actually indicate ``predicate'', not ``verb'', but we use ``verb'' to indicate the grammatical element in this work.}

We use a very simple set of 12 alignment constraints, which are also sometimes called \emph{edge-most} constraints \citep{Prince1993, prince2004, mccarthy03}. In this work, the alignment constraints are a binary, not distance-based, indication of whether or not an element is aligned to the edge of a sentence. Our constraints are similar to those used by \cite{Costa97,Costa}, who used an OT framework to explain variation in unmarked word orders over many languages. Unlike that work, we use alignment constraints on grammatical structures in addition to information structure markers. Six of our constraints are grammatical structure alignment constraints corresponding to the left and right alignment of the subject, verb, and object (e.g. subject left). Similarly, six are information structure alignment constraints corresponding to the left and right alignment of each of the information structure markers enumerated in Section \ref{sec:data}. All alignment constraints are evaluated with respect to entire sentences, not, for example, with respect to the verb phrase only. Specifically, these universal alignment constraints are:

\begin{itemize}
\item Grammatical structure alignment constraints:
\begin{itemize}
\item{Align Subject Left}
\item{Align Subject Right}
\item{Align Verb Left}
\item{Align Verb Right}
\item{Align Object Left}
\item{Align Object Right}
\end{itemize}
\item Information structure alignment constraints:
\begin{itemize}
\item{Align C-Topic Left}
\item{Align C-Topic Right}
\item{Align Topic Left}
\item{Align Topic Right}
\item{Align Focus Left}
\item{Align Focus Right}
\end{itemize}
\end{itemize} 

Formally, the Align C-Topic Left constraint is given by:
\begin{enumerate}
\item{Align C-Topic Left: Contrastive topics are sentence initial. Failed when a contrastive topic exists in the sentence and a contrastive topic is not sentence initial. Vacuously satisfied if a contrastive topic is not present in the sentence.}
\end{enumerate}
The formal definitions of the other alignment-based constraints follow directly from this.

Each of the 12 constraints is violated when the surface word order does not obey the stated left or right alignment. For example, an SVO surface word order violates the constraints {\it align subject right}, {\it align verb left}, {\it align verb right}, and {\it align object left}. Also for an SVO surface word order, if the subject is marked as a contrastive topic and the verb and object are both marked as focused elements, then the constraints {\it align c-topic right} and {\it align focus left} are violated. The constraints {\it align topic left} and {\it align topic right} are vacuously satisfied as there is no topicalized element in the example sentence.

\subsection{Data Variation}\label{sec:dvar}

The sentences and corresponding annotations that we extract from the PDT include variation due to optionality and also variation due to incomplete information, or sentence representations. The machine learning community refers to this type of variation, legitimate or not, as noise in the data. The learning algorithms that we use to automatically make sense of the data must be tolerant of these types of variation, or noise. %We assume that there is some noise, or mistakes, in the data. These may be a result of annotation errors or errors in our data preprocessing, %\footnote{Although we seek to identify the simple, canonical sentences in the PDT, we may include some sentences with more complex structure than we intend or assume to be learning from.}
%but we expect both to be infrequent. Additionally, 

First, the data includes evidence of true variation, or grammatical word order optionality. That is, for some meaningful content situated in a particular discourse context, there may be more than one grammatical word order (e.g. speakers may have the option of using {\it SVO} or {\it SOV} order). Second, some of the observed variation in the data may be due to grammar or information structure that we do not consider. That is, if we considered additional properties of a given sentence, it may be possible to explain away why one word order is observed instead of another. Teasing apart true variation and variation due to incomplete information would require having native Czech speakers closely examine each sentence in our dataset as well as their discourse context (preceding sentences) and give detailed additional annotations, possibly beyond those represented in the PDT. %Instead, we assume that both exist and attempt to learn as much as we can from the data and annotations that we have. 
In this section, we present the data and a discussion of the variation that we observe in it. Furthermore, we describe lower and upper bounds on word order prediction accuracy, given the input to the learners, which determine what we expect and hope for our learning algorithms to achieve on similarly behaved data.

Table \ref{table:bigdv} shows input feature patterns (all combinations of subject, verb, and object marked by topic, contrastive topic, or focus) and the frequencies of observed surface word orders. This data provides an upper bound on how well we can expect learning algorithms to be able to predict word order, given only the information structure of the subject, verb, and object. There are 27 input patterns (each of S, V, and O marked with either T, C, or F). Given only this information, the best any learner can do is to produce the most likely word order for each input pattern, which would achieve 68.1\% accuracy on the training data. %It is interesting to note that most of the word order variation is captured by just two word order choices (of the possible six) for each of the input patterns. That is, if a perfect learner is allowed to produce two possible word orders for each input pattern, one of those word orders would be correct for 90.1\% of the observed sentences in the training data. 
In Section \ref{sec:resultsWO}, we will say that the prediction strategy of memorizing the single best word order for each of the 27 input patterns is the upper limit in model performance for predicting the word order of a sentence. This strategy is the upper bound because no additional information about the sentence will be available to the learner.

Table \ref{table:bigdv} also shows that, given no information about the information structure of the elements in a sentence, the best strategy for predicting word order is to, simply, always predict SVO. This approach would correctly predict the word order of 50.0\% of the sentences represented in the table. %Similarly, if we are able to predict two word orders, we should always predict SVO and OVS, the two most common of the six. Using this strategy, the correct word order would be one of the two predictions for 68.3\% of the sentences. 
In Section \ref{sec:resultsWO}, we will refer to this as the baseline strategy for predicting the word order of a sentence. 

\begin{figure}[h!]
\vskip 0.0in
\begin{center}
\includegraphics[width=1.0 \linewidth]{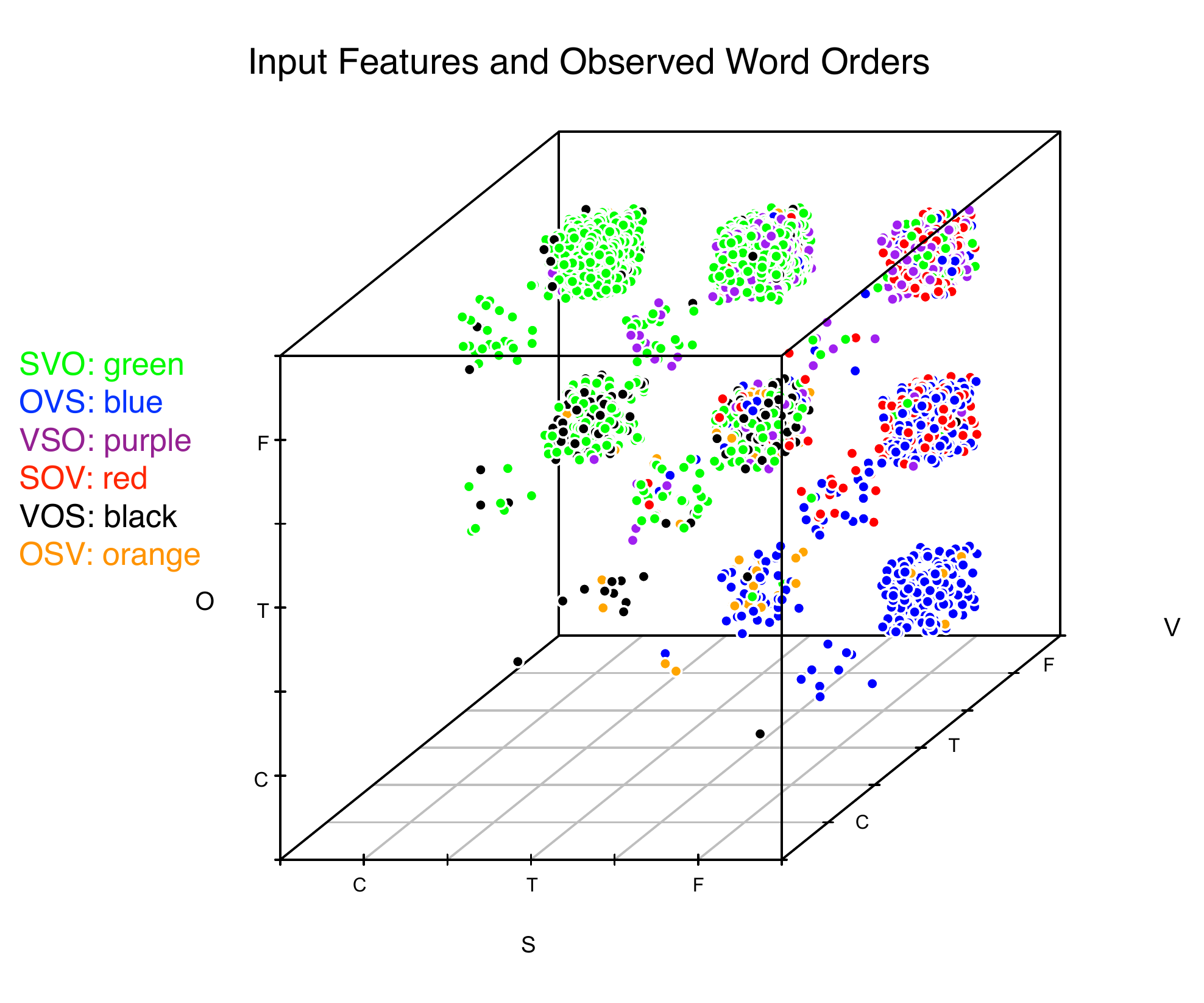}
\vskip -0.15in
\caption{Visualization of input sentences: The alignment of S, V, and O with C, T, and F is plotted, and the observed word orders are indicated by colored points. Each cluster represents the set of sentences with a particular input pattern, and positions within clusters are meaningless.}
\label{fig:colorviz} 
\end{center}
\vskip -0.2in
\end{figure}

Figure \ref{fig:colorviz} is a visualization of the data in Table \ref{table:bigdv}. The 27 input patterns are shown in clusters on a three dimensional plot. The axes correspond to grammatical structure (S, V, and O) and values along each axis correspond to the information structure associated with each for a particular sentence in the training data. Colors indicate the word order of the observed sentences. It is easy to see that most clusters are dominated by one or two word orders and that some input patterns (like V-C) rarely, if ever, occur in the data. Finally, many of the clusters are dominated by either SVO or OVS word orders, but there is little overlap of the two within a single input pattern. SVO more often occurs with input patterns that also yield VOS while OVS more often occurs with input patterns that also yield VSO and SOV. 

\begin{table}[h!]
\begin{center}
\small
\begin{tabular}{|c|c|c||c|c|c|c|c|c|c||c|}
\hline
\multicolumn{3}{|c||}{Disc Func} & \multicolumn{7}{|c||}{Counts of Sentences with Basic Word Orders} & \% with Most \\
\cline{1-10}
S & V & O & SVO & OVS & VSO & SOV & VOS & OSV& Sum & Likely Order \\
\hline
\hline
T & T & T & {\bf 23} & 4 & 4 & 3 & 3 & 3 & 40 & 58\% \\ %& 68\% \\
T & T & C & 0 & 1 & 0 & 0 & 0 & {\bf 2} & 3 & 67\% \\ % & 100\% \\
T & T & F & {\bf 22} & 0 & 11 & 0 & 1 & 0 & 34 & 65\% \\ % & 97\% \\
T & C & T & 0 & 0 & 0 & 0 & 0 & 0 & 0 & - \\ %& - \\
T & C & C & 0 & 0 & 0 & 0 & 0 & 0 & 0 & - \\ %& - \\ 
T & C & F & 0 & 0 & 0 & 0 & 0 & 0 & 0 & - \\ %& - \\
T & F & T & {\bf 97} & 26 & 28 & 12 & 80 & 32 & 275 & 35\% \\ %& 64\% \\
T & F & C & 2 & {\bf 43} & 0 & 0 & 1 & 20 & 66 & 65\% \\ %& 95\% \\
T & F & F & {\bf 519} & 7 & 145 & 17 & 28 & 4 & 720 & 72\% \\ %& 92\% \\
C & T & T & {\bf 7} & 0 & 0 & 0 & 3 & 0 & 10 & 70\% \\ %& 100\% \\
C & T & C & 0 & 0 & 0 & 0 & {\bf 1} & 0 & 1 & 100\% \\ %& 100\% \\
C & T & F & {\bf 26} & 0 & 1 & 0 & 3 & 0 & 30 & 87\% \\ %& 97\% \\
C & C & T & 0 & 0 & 0 & 0 & 0 & 0 & 0 & - \\ %& - \\
C & C & C & 0 & 0 & 0 & 0 & 0 & 0 & 0 & - \\ %& - \\
C & C & F & 0 & 0 & 0 & 0 & 0 & 0 & 0 & - \\ %& - \\
C & F & T & {\bf 111} & 0 & 2 & 0 & 76 & 4 & 193 & 58\% \\ %& 97\% \\
C & F & C & 0 & 0 & 0 & 0 & {\bf 9} & 2 & 11 & 82\% \\ %& 100\% \\
C & F & F & {\bf 610} & 0 & 3 & 0 & 34 & 2 & 649 & 94\% \\ %& 99\% \\
F & T & T & 1 & {\bf 17} & 1 & 14 & 0 & 0 & 33 & 52\% \\ %& 94\% \\
F & T & C & 0 & {\bf 9} & 0 & 0 & 0 & 0 & 9 & 100\% \\ %& 100\% \\
F & T & F & 4 & 3 & {\bf 5} & 2 & 0 & 0 & 14 & 36\% \\ %& 64\% \\
F & C & T & 0 & 0 & 0 & 0 & 0 & 0 & 0 & - \\ %& - \\
F & C & C & 0 & 0 & 0 & 0 & {\bf 1} & 0 & 1 & 100\% \\ %& 100\% \\
F & C & F & 0 & 0 & 0 & 0 & 0 & 0 & 0 & - \\ %& - \\
F & F & T & 7 & {\bf 222} & 16 & 153 & 4 & 2 & 404 & 55\% \\ %& 93\% \\
F & F & C & 0 & {\bf 184} & 0 & 1 & 0 & 4 & 189 & 97\% \\ %& 99\% \\
F & F & F & 48 & 24 & {\bf 105} & 95 & 1 & 0 & 273 & 38\% \\ %& 73\% \\
\hline
\multicolumn{3}{|l||}{\multirow{2}{*}{Totals}} & 1477 & 540 & 321 & 297 & 245 & 75 & 2955 & 2013 \\ %& 2661 \\
\multicolumn{3}{|l||}{} &  50.0\% & 18.3\% & 10.9\% & 10.1\% & 8.3\% & 2.5\% & & 68.1\% \\ % & 90.1\% \\
\hline
\end{tabular}
\caption{Input patterns of the alignment between information structure and grammatical structure and counts of basic word order frequencies in training set of 2955 PDT sentences. The frequency of the most common word order for each input is bolded. Not all inputs are observed in the data.}%, and the percent of all sentences with the given input pattern that have the most frequent word order is given in the last column. } %Given this input, 68.1\% is the upper bound in performance for predicting the most likely word order. }
\label{table:bigdv}
\end{center}
\end{table}

It is important to recall here that the input information shown in Table \ref{table:bigdv} and Figure \ref{fig:colorviz} (e.g. the verb is focused, or the subject is topicalized) \emph{is not} directly used in learning. Rather, this information, in combination with a hypothesis (a word order), determine which \emph{constraints} are violated. The input to the grammar consists of the constraint violations for a given candidate word order. In Section \ref{sec:learning}, we show how weights corresponding to constraints are learned and then used to predict labels on new data. 

In addition to predicting the single most likely word order for a given pattern of input, we would like the grammars that our algorithms learn to be able to predict the observed variation in word order outputs \citep{Bresnan07, Doyle08}. That is, for example, we would like learned grammars to output SVO as the most likely word order for a topicalized subject and a focused verb and object and, additionally, output some probability distribution over the output word orders that indicates that VSO also has a fairly high probability, though it is not as high as that of SVO (see Table \ref{table:bigdv}).

\section{Learning}\label{sec:learning}
\subsection{Perceptron Algorithm}
The perceptron algorithm was introduced by \cite{rosenblatt1958} as a theory of the way a hypothetical biological nervous system stores and learns information. Since then, the algorithm has become a popular method in machine learning for doing supervised, discriminative classification \citep{Littlestone:1988} and there is considerable research expanding Rosenblatt's original proposal (e.g., \cite{Collins:2002, Khardon05}). 

Supervised learning algorithms like the perceptron learn to make predictions, such as a label in the case of discriminative classification, about input data given training examples that contain both input data and the true value of some underlying form, such as a label. In this work, we train a perceptron discriminative classifier, which chooses among the six possible sentential word orders ({\it SVO}, {\it SOV}, etc.). Training data takes the form of constraint violations determined by information structure markers on grammatical elements (e.g. {\it subject is topicalized}), the input data, and the corresponding observed word order pattern (the label). Using a supervised learning algorithm like the perceptron is a natural way to model human language learning since human learners also learn from observing labeled examples (individual sentences are supervised examples, and humans observe their word order).

The perceptron is an online algorithm, i.e. it updates its parameters based on one example (in this case, one sentence) at a time. Its only learned parameter is a single list of weights, each of which is associated with one attribute\footnote{The machine learning community refers to these as {\it features}. For clarity in this linguistic context, we refer to them as attributes.}. Attributes indicate the value of some property of a single input example. In this work, attributes correspond to HG/OT constraints and the perceptron learns weights associated with each. A hypothesis (word order) is scored by computing a linear combination of attribute values (given the input and hypothesized label) and the corresponding weights.\footnote{For example, say the model only contains two attributes, A and B, and the current model weight vector gives A a weight of 3 and B a weight of 5, or $\vec{w}=<3, 5>$. If some input and hypothesis has attribute values of $f_{A}(input, word order) = -1$ and $f_{B}(input, word order) = 1$, and $\vec{f}_{AB}(input, word order) = <-1,1>$, then the resulting score would be ${\vec{f}_{AB}}*\vec{w}^{T}$ , or $-1*3 + 1*5 = 2$.} Following \cite{boersmaPater08}, in this work, all attribute values are one of the following: $1$, indicating that the input and hypothesis explicitly complies with a constraint; $0$, indicating that a constraint is vacuously satisfied; or $-1$, indicating the input and hypothesis violate a constraint. That is, the harmony, or linear combination of weights, of a hypothesis is a function of the constraints that it complies with as well as those that it violates. \cite{boersmaPater08} point out that this setup is not explicitly disallowed by the original formulation of OT, which did not distinguish between overt constraint satisfaction and vacuous constraint satisfaction.

The perceptron's learning is mistake-driven. That is, it only updates its parameters when it makes a mistake in predicting the label of an item in the training set. It iterates through the training data one example at a time, predicts the label of each example given its current parameters, and, finally, if its prediction doesn't match the true label, it updates its weight vector accordingly. 

Specifically, the prediction function is:
$$\hat{y}_{i}=\argmax_{y \in Y}  f(x_i,y) \cdot w $$
where $\hat{y}_i$ is the predicted label, for input $x_i$, which has the highest score among all possible labels $Y$. Scores, or, in our case, harmonies, are calculated by multiplying the attribute vector, $f(x_i,y)$, by a vector of the current weights, $w$. The $\cdot$ indicates the dot product between the two vectors. As mentioned, in this work, each attribute has a value of $1$ (the constraint associated with the attribute is complied with, given the input $x$ and the hypothesis  $y$), $0$ (the constraint is vacuously satisfied), or $-1$ (the constraint associated with the attribute is violated).\footnote{This setup could easily be translated into one where all attributes have a value of $1$ or $0$ by doubling the attribute set to include one attribute for complying with each constraint and one attribute for violating each constraint. Appendix \ref{sec:learningex} explains the $1$/$-1$ attributes in detail.} The update function for learning is as follows:

\vspace{5 mm}
\noindent{}for each training example $x_i$:

%\item if $ \hat{y_i} \neq y_i:$
$ w_{t+1} = w_t + \lambda(f(x_i,y_i)-f(x_i, \hat{y_i}))$

\vspace{5 mm}

\noindent{}where $y_i$ is the true label, $\hat{y}_i$ is the predicted label, $w_t$ is the current weight vector, $w_{t+1}$ is the updated weight vector, and $\lambda$ is a learning rate parameter. The difference between the attribute values for the true label and for the predicted label ($f(x,y)-f(x,\hat{y})$) is the set of attributes (constraints) over which the true and predicted labels differ. If the true and the predicted labels are the same, the difference is zero and no update is made. 

Appendix \ref{sec:learningex} gives a detailed example of how the perceptron algorithm learns a weight vector corresponding to the twelve constraints listed in Section \ref{sec:constraint} from the PDT data described by Table \ref{table:bigdv}.

\subsection{Comparison with Other OT Learners}\label{sec:learningcomp}
In this section, we give intuitive, theoretical comparisons of the perceptron learner with algorithms used in previous OT/HG learnability research: the Constraint Demotion (CD) algorithm, the Gradual Learning Algorithm (GLA), and the Maximum Entropy (MaxEnt) classifier. Table \ref{table:summarycomp} summarizes the discussion.

Constraint Demotion (CD) was the first OT learning algorithm to be proposed \citep{Tesar1998, Tesar2000}. The CD algorithm assumes that there always exists a discrete, hierarchical constraint ranking, and the ranking is updated when the learner observes a piece of evidence for which she would not have produced the proper surface form, given the input data. Say the winner is the true, observed surface form and the loser is the surface form that the current model incorrectly predicted. The update function demotes all of the constraints that the winner violated which are currently ranked higher than the highest ranked loser's constraint violation. At the end of the update, the learner is guaranteed to correctly predict the true, observed surface form. This algorithm, like the perceptron, is online and mistake-driven. However, it is not at all robust to data variation and will result in endless rank swapping if multiple surface forms for a single input are observed. \cite{Tesar2000} discuss this limitation in detail, and it is the motivation for the Gradual Learning Algorithm, which was proposed soon after CD.

\cite{Boersma1997} is the first presentation of the Gradual Learning Algorithm (GLA), which is an extension of the CD algorithm aimed at making it more robust to data variation. In the GLA, instead of aggressively demoting constraint rankings, numerical weights are associated with each and these are gradually updated\footnote{The size of the updates is a model parameter. In the perceptron formulation, this value is generally referred to as the learning rate. In the original GLA proposal, it is called the plasticity.} when the learner makes mistakes. Thus, the GLA updates are similar to the perceptron algorithm's updates. However, at the time of prediction, the GLA uses the magnitude of the weights to determine a discrete, non-numerical OT constraint ranking, and it uses the OT ranking to predict surface forms. 
Additionally, in order to account for data variation, the GLA may learn a {\it Stochastic Optimality Theory} (SOT) grammar rather than a traditional OT grammar. 
An SOT grammar adds some random noise\footnote{The noise is sampled from a zero mean, standard deviation one Gaussian and multiplied by a model parameter, called the spreading value in the original proposal of the GLA} to the values of each constraint before extracting an OT constraint ranking and predicting the surface form. Like the perceptron and CD, the GLA is online and mistake-driven. However, it is different from the perceptron in that it uses an OT or SOT constraint ranking moel for predicting surface forms, not allowing for so-called ganging-up-effects.

%\cite{boersmaPater08} extend this idea and propose a Harmonic Grammar version of the GLA algorithm (HG-GLA), which is similar to the perceptron algorithm. The HG-GLA work provides a detailed theoretical exploration of the learning algorithm, including some learning computer simulations but no empirical investigation. The authors point out that HG is a promising model of generative grammar. We explore each of these algorithms in more detail in Section \ref{sec:learningcomp}.

Recently, \cite{boersmaPater08} proposed a Harmonic Grammar version of the GLA, HG-GLA. This algorithm learns just like the perceptron and the authors replicate perceptron proofs of convergence to explain some of the theoretical properties of the HG-GLA. Because that work is preliminary and only includes computer simulations, we do not adopt its terminology. % and, rather, use the long-standing perceptron formulation and terminology.

Maximum Entropy (MaxEnt) models, or log-linear models, %\footnote{the type used in OT/HG grammar learnability are also known as multinomial logistic regression} 
are frequently used in the Machine Learning community for many classification tasks. Like the perceptron, MaxEnt models learn a vector of weights associated with each attribute function defined for the dataset (in our case, 12 constraints). However, MaxEnt learners seek to mimic training data distributions (while otherwise assuming as much variance, or uniformity, in the model as possible) and are designed to, by exponentiating and normalizing the linear combination of weights and attributes for a given input, produce a probability distribution over output labels, or surface word orders, in our case. Therefore, unlike the CD, perceptron, and GLA, the output of a MaxEnt classifier is a probability distribution over outputs. The most probable label, or surface form, is the maximum likelihood prediction for a given input. In contrast, the other three algorithms most naturally produce only a single maximum likelihood output prediction, and we must do some extra work\footnote{Such as inserting some random variation and sampling from outputs.} to produce a distribution over output labels.

MaxEnt classifiers are typically learned in batch mode, not in an online setting like CD, the GLA, or the perceptron. That is, the model parameters are estimated by looking at all of the training data at once, not one sentence at a time. However, unlike CD and the GLA but like the perceptron, the MaxEnt model allows for violated low ranked constraints to collectively outrank a violation of a higher ranked constraint. \cite{Goldwater2003} use a MaxEnt model to learn probabilistic phonological grammars both with and without variation. That work uses a MaxEnt model to explain variation in Finnish genitive plural endings as accurately as the GLA model does and it claims that the model is more theoretically sound than the GLA. \cite{Jaeger2003} use a MaxEnt model to predict variation in the English genitive construction\footnote{e.g.  \emph{the mother's future} vs. \emph{the future of the mother}} and provide some evidence that it is superior to the GLA model because of its ability to allow for ganging-up effects of low ranked constraints. Finally, \cite{HayesWilson08} use a MaxEnt model to learn a model of phonotactic patterns and \cite{pater10} explore the model's convergence properties.

%the animacy, topicality, and possessive attributes of the noun phrases involved. Unlike the previous work in OT grammar learnability, \cite{Jaeger2003} learn syntactic constructions. However, their dataset is unique in that it was gathered from an experimental questionnaire testing English speakers' preferences. Additionally, their data is concerned with preferences among grammatical constructions, rather than grammaticality itself.

In this work, we claim that the perceptron has the benefit of being an online (one sentence at a time) learner, like CD and the GLA, but that it is able to outperform those algorithms because it uses the HG formalism and allows for ganging-up effects, like MaxEnt models. Although it does not naturally produce probability distributions over output labels, like MaxEnt models, we show in Section \ref{sec:varresults} that it does a good job of predicting variation. Additionally, it has the advantage of being a simple but well-established and well-studied learning algorithm.

\begin{table}
\begin{center}
\begin{tabular}{|l|c|c|c|}
\hline
Learner & OT/HG-Specific & Online & Prob Dist Output\\
\hline
Constraint Demotion  & Yes & Yes & No \\
Gradual Learning Alg.  & Yes & Yes & No \\ 
Perceptron  & No & Yes & No \\
Maximum Entropy  & No & No & Yes \\
\hline
\end{tabular}
\caption{Summary of the discussion in Section \ref{sec:learningcomp}. Yes/No values represent the most natural properties of each learner. It may be possible to, for example, output a probability distribution over outputs from GLA and perceptron outputs or learn the parameters of a MaxEnt model using an online version of a learning algorithm. The OT/HG-Specific property indicates whether the learning algorithm was created specifically to learn an OT or HG grammar or not. Because the perceptron and MaxEnt learners have been widely used in many fields, they are well-studied and their properties and well-known.}
\label{table:summarycomp}
\end{center}
\end{table}

\subsection{Data Setup}
For learning, the set of 4955 simple sentences extracted from the PDT are split into training, development, and test sets. The parameters (constraint weights) for a Harmonic Grammar are learned from the 2955 training data sentences. The 1000 development set sentences are used to choose a specific learning model (we explore several variants of the perceptron learning algorithm), and the remaining held-out 1000 test set sentences are used for evaluation after we identify a final model and parameters. 

The subject, object, and predicate in each of the sentences is annotated as topicalized, contrastively topicalized, or focused. The attribute functions use these annotations and the hypothesis (a word order) to determine whether or not each constraint is violated (i.e. whether each of the 12 attributes has a value of $1$, $0$, or $-1$ for some input and word order). As usual, a single weight vector is learned from the entire dataset. The constraint set and attribute functions are identical for learning GLA, perceptron, and MaxEnt models.

\subsection{Perceptron Limitations and Modifications}
The perceptron has several well-known limitations. These include slow convergence time and an inability to account for unbalanced data (in terms of attribute values and/or output labels). In order to overcome these limitations, we make some standard slight modifications to the algorithm and choose our final learning setup by measuring performance on the 1,000 sentence development set.

\begin{figure}
\vskip 0.0in
\begin{center}
\includegraphics[width=0.8 \linewidth]{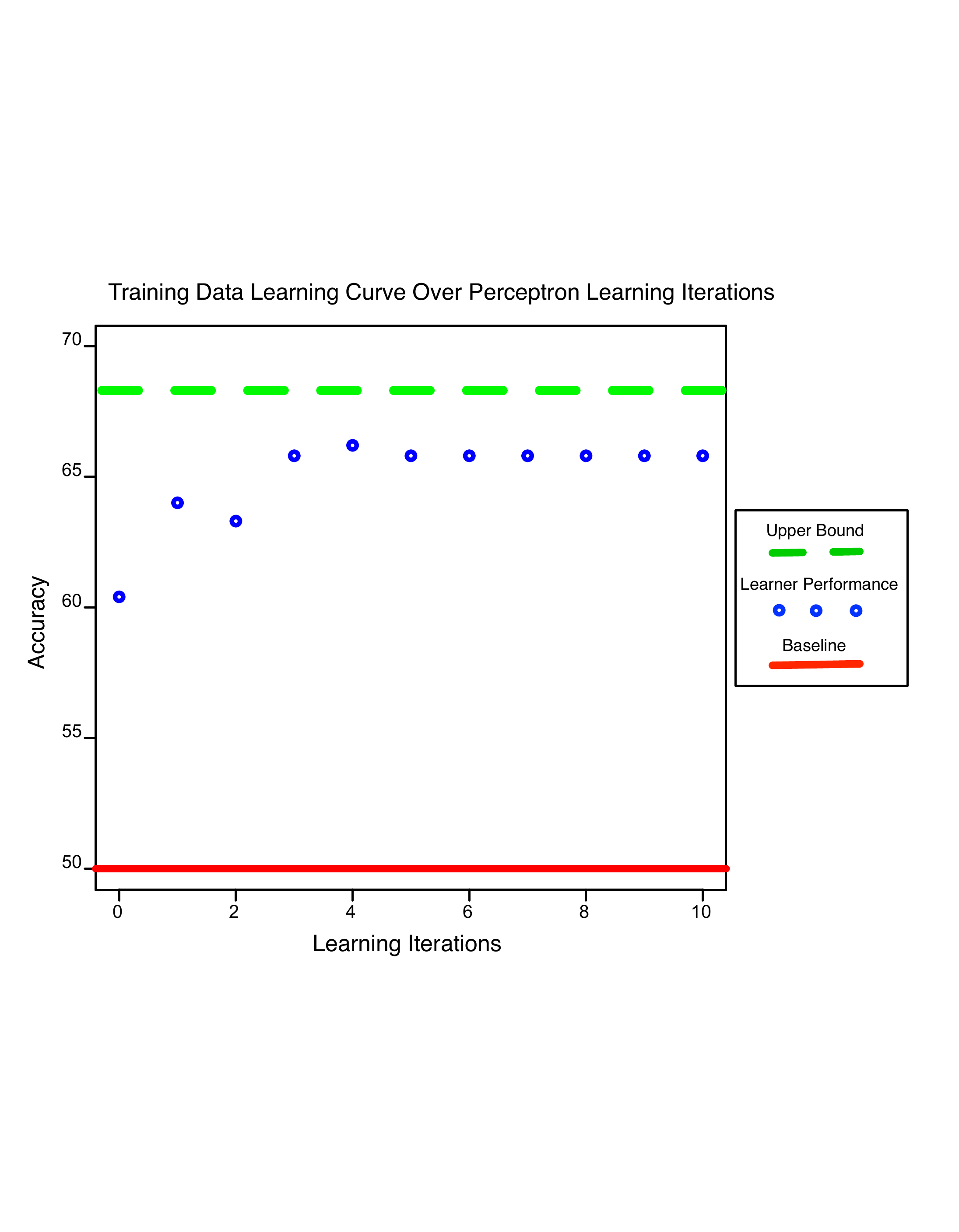}
\vskip -1.2in
\caption{Learning curve over many iterations over the training data. The baseline performance is shown in a red solid line and the upper bound performance in green dashes. Our learner's accuracy over ten iterations is shown in blue circles. Learning converges after just a few iterations over the training data.}
\label{fig:lcTrainIt} 
\end{center}
\vskip -0.4in
\end{figure}

{\bf Learning iterations}
Because the perceptron converges slowly, we have it learn over many iterations over the training data. This is standard practice in the machine learning community, where the availability of supervised data is often an issue. It is also a reasonable strategy from the viewpoint of human learnability. Our data sample really contains sentence {\it types}, rather than sentence {\it tokens}. Our learner does not have access to the lexical items in a given instance in the training data (e.g. every sentence of type Subject-T, Verb-F, Object-T looks the same to the learner, no matter the actual words in the sentence). Therefore, as long as our sample of input types is representative and we can assume that a human learner would observe a constant distribution of input types over time (e.g. a child observes the same frequency of Subject-T, Verb-F, Object-T sentences in her first month as she does in the second month), then iterating over a single set of types is a reasonable approach. 
%is very small in comparison to real language learning situations. %Furthermore, it is reasonable to assume that human learners may be exposed to single sentences multiple times. 
However, we do find that learning converges over just a few iterations over the training data, as shown in Figure \ref{fig:lcTrainIt}. %Figure \ref{fig:lcSentIt} shows a learning curve over sentences in several iterations over the training data. [Note: maybe get rid of upper bound and baselines from these graphs, they don't add anything]

{\bf Learning in the presence of noise}
Although the perceptron is fairly tolerant to noise (in our case, true variation in the data), it is not completely immune. In order to increase its resistance, we implement a slight modification to the original algorithm. The $\lambda$-trick, explained in \cite{Khardon05}, allows the algorithm to ignore training examples that it repeatedly misclassifies, that is, those examples which look like they contain noise. In other words, it helps the algorithm learn to ignore sentences that look very different from the rest of the data. Predictions are made as follows:

$$\hat{y}_{i}=\argmax_{y \in Y}  [ f(x_i,y) \cdot w  + \lambda I_{y_i}C_{i}] $$
where $I_{y_i}$ is an indicator function that is 1 if $y$ is the correct label $y_i$ and 0 otherwise. $C_{i}$ is a counter that gives the total number of times example $i$ has been incorrectly classified and $\lambda$ is a learning parameter. So, as $C_i$ increases, $\lambda I_{y_i}C_i$ will eventually dominate the sum for the correct label $y_i$, resulting in the correct prediction and no update. As before, $\hat{y}_i$ is the predicted label, for input $x_i$, which has the highest score among all possible labels $Y$. Scores, or, in our case, harmonies, are calculated by multiplying the attribute vector, $f(x_i,y)$ by a vector of the current weights, $w$. It is important to keep in mind that this `trick' is only used in {\it training}, of course. 
%The $\cdot$ indicates the dot product between the two vectors. As mentioned, in this work, each attribute has a value of $1$ (the constraint associated with the attribute is complied with, given the input $x$ and the hypothesis  $y$), $0$ (the constraint is vacuously satisfied), or $-1$ (the constraint 

Again, this modification is reasonable from the perspective of human language acquisition; there may be factors that would enable a human learner to recognize which sentences are different from the rest (e.g. spoken emphasis) or, simply, a sentence may be different enough from the others that the listener assumes it is wrong. 

{\bf Update normalization}
Finally, the perceptron is not good at learning when the training data is not well balanced with respect to attributes or output labels, and the PDT dataset has both types of unbalances. In order to account for this limitation, we normalize the update values using counts of the number of times a given attribute is either directly complied with or violated, but not vacuously satisfied, in the training data. For example, there are many sentences in the training data that have no contrastive-topical element. For those sentences, both {\it align c-topic right} and {\it align c-topic left} are vacuously satisfied. Therefore, the number of times that these attributes directly apply to a sentence is less than, for example, {\it align subject left}, which applies to all sentences because every sentence has a subject. When an update is made to an attribute, %we divide the learning rate parameter, $\eta$, 
we divide by the normalization factor for the attribute. As a result, the size of the updates made to {\it align c-topic right} is greater than, for example, the size of the updates made to {\it align subject left}. This is a reasonable normalization technique since we want less frequent attributes to have the chance to achieve the same constraint weight magnitude as more frequent attributes. 

%For purposes of reproducibility and encouraging future research, we plan to make our data and code available to the research community. 
 
%\begin{figure}[h!]
%\vskip 0.0in
%\begin{center}
%\includegraphics[width=0.8 \linewidth]{figures/lc-train-sent-plotted.pdf}
%\vskip -0.15in
%\caption{Learning curve over sentences in several copies of the training data.Top-1 and top-2 baselines are shown in black, top-1 and top-2 upper bounds are shown in pink, our top-1 is shown in blue, and our top-2 is shown in green. All top-1 accuracies are indicated with circles and top-2 accuracies are indicated with triangles.}
%\label{fig:lcSentIt} 
%\end{center}
%\vskip -0.2in
%\end{figure}

\section{Results}\label{sec:results}

\subsection{Perceptron Word Order Prediction Accuracy}\label{sec:resultsWO}

In Section \ref{sec:dvar}, we described the baseline prediction strategy (always predict SVO) as well as the upper bound (consider the complete input and predict the most probable label) for predicting the word order of a Czech sentence given the information available to our learning algorithm. Accuracy is defined as the percent of test set inputs for which a given model correctly predicts the surface word order. Table \ref{table:results} shows these accuracies on a held out test set as well as the accuracy of the model learned by the perceptron algorithm. As the results show, the perceptron's model drastically outperforms that baseline and its accuracy is within 2.5\% of the upper bound.

\begin{table}[h]
\begin{center}
\begin{tabular}{|l|c|}
\hline
& Accuracy \\ %& Top-2 Accuracy \\
\hline
Baseline & 50.3\% \\ % & 68.6\% \\
Perceptron Performance & 67.0\% \\ %& 78.3\%\\
Upper Bound & 69.4\% \\ % & 91.5\%\\
\hline
\end{tabular}
\caption{Baseline, perceptron, and upper bound prediction strategy accuracies on a held out set of 1,000 test sentences. Because weights are initialized randomly, there is some minor variation in the accuracy scores. Perceptron performance results are averaged over ten runs, and the standard deviation is 0.15\%.}
\label{table:results}
\end{center}
\end{table}

\subsection{Harmonic Grammar Weights}
Three sets of constraint weights learned by three randomly initialized runs of the perceptron learning algorithm are shown in Table \ref{table:weights}. It is interesting to see that the weights as well as the overall relative rankings do vary between the runs. However, the right/left preference for each information structure element (C-Topic, Topic, Focus) and two of the grammatical structure elements (Subject, Object) does \emph{not} vary. Table \ref{table:weightsdiffs} shows the difference between the learned weight for the left alignment constraint and the learned weight for the right alignment constraint for each element. All runs prefer C-Topic to be left, Topic to be left, and Focus to be right. Similarly, all runs prefer the object to be right and, perhaps contrary to expectations because the most common Czech word order is SVO, the subject to be right. The only element for which the learners do not learn a consistent alignment preference is the verb. However, all weights learned for the verb constraints are relatively small. In fact, this indicates that the model prefers the the verb to be neither left nor right but, rather, in the middle. That is, the when the verb is in the middle of a sentence, both of its alignment-based constraints are violated but neither penalty (constraint weight) is large. This is consistent with the fact that the two most frequent word orders are SVO and OVS.

%Table \ref{table:weights2} shows the total weight magnitude (for both left and right alignment constraints) for the same three runs shown in Table \ref{table:weights} for each of the six grammatical and information structure elements. With one exception (Run 2, Topic) each run assigns the same relative importance to the alignment of the six elements. In order from most to least important, they are: C-Topic, Topic, Focus, Object, Subject, Verb. It is interesting to note that the alignments of each of the three information structure elements are all more important than the alignment of any of the grammatical structure elements. 

Additionally, Tables \ref{table:weights} and \ref{table:weightsdiffs} clearly show that the learned constraint weights have the same order of magnitude. That is, the weights of the lower ranked constraints are not so much less than those of the higher ranked constraints that we would never expect them to influence the word order output. In contrast, the relative magnitude of the lower ranked constraints indicate that so-called ganging-up-effects may be important and occur somewhat frequently. Below, in Section \ref{sec:glaresults}, we compare the perceptron with the GLA and empirically show that allowing for such effects leads to better word order prediction accuracies.

\begin{table}
\begin{center}
\begin{tabular}{|l|c|c|c||c|c|c|}
\hline
\multirow{2}{*}{Constraint} & \multicolumn{3}{c||}{Learned Weight} & \multicolumn{3}{|c|}{Learned Ranking}\\
 & \multicolumn{1}{c}{Run 1} &  \multicolumn{1}{c}{Run 2} &  \multicolumn{1}{c||}{Run 3} & \multicolumn{1}{c}{Run 1}&  \multicolumn{1}{c}{Run 2} &  \multicolumn{1}{c|}{Run 3}\\
\hline
C-Topic Left & 15.60 & 12.28 & 12.95 & 1 & 2 & 1 \\
Topic Left & 11.26 & 8.79 & 10.33 & 2 & 5 & 3 \\
Focus Right & 10.39 & 10.12 & 9.57 & 3 & 4 & 4 \\
Topic Right & 9.36 & 8.40 & 8.42 & 4 & 6 & 7 \\
Object Right & 8.63 & 13.29 & 10.41 & 5 & 1 & 2\\
Focus Left & 8.36 & 7.79 & 7.98 & 6 & 8 & 8 \\
C-Topic Right & 7.21 & 7.34 & 8.69 & 7 & 9 & 6 \\
Subject Right & 7.18 & 12.01 & 9.01 & 8 & 3 & 5 \\
Object Left & 6.99 & 4.30 & 5.49 & 9 & 10 & 11 \\
Subject Left & 5.96 & 4.21 & 5.64 & 10 & 11 & 10 \\
Verb Left & 5.68 & 3.52 & 4.88 & 11 & 12 & 12 \\
Verb Right & 3.40 & 7.95 & 6.62 & 12 & 7 & 9 \\
\hline
Accuracy & 67.0\% & 67.0\% & 67.0\% \\
\cline{0-3}
\end{tabular}
\caption{Weights learned by the perceptron algorithm. The three sets of weights shown here correspond to three of the ten runs averaged in Table \ref{table:results}. Constraint weights as well as the overall relative ranking vary some over the individual runs, but the relative ranking of the left and right alignment constraints for individual elements tends to be consistent.}
\label{table:weights}
\end{center}
\end{table}

\begin{table}
\begin{center}
\begin{tabular}{|l|p{2.cm}|p{2.cm}|p{2.cm}||p{2.cm}|}
\hline
\multirow{2}{*}{Element} & \multicolumn{4}{c|}{Difference Between Left and Right Aligned Constraint Weights, $W_{left} - W_{right}$}\\
& \multicolumn{1}{c}{Run 1} &  \multicolumn{1}{c}{Run 2} & \multicolumn{1}{c||}{Run 3}  & \multicolumn{1}{c|}{Average} \\
\hline
C-Topic & \hspace {0.7cm} 8.39  & \hspace {0.7cm} 4.94 & \hspace {0.7cm}  4.26 & \hspace {0.7cm} 5.86\\
Object & \hspace {0.65cm} -1.64  & \hspace {0.65cm} -8.99 & \hspace {0.65cm}  -4.92 &  \hspace {0.65cm} -5.18\\
Subject & \hspace {0.65cm} -1.22  & \hspace {0.65cm} -7.80 & \hspace {0.65cm}  -3.37 & \hspace {0.65cm} -4.13\\
Focus & \hspace {0.65cm} -2.03  & \hspace {0.65cm} -2.33 & \hspace {0.65cm}  -1.59 & \hspace {0.65cm} -1.98 \\
Topic & \hspace {0.7cm} 1.90  & \hspace {0.7cm} 0.39 & \hspace {0.7cm}  1.91 & \hspace {0.7cm} 1.40\\
Verb & \hspace {0.7cm} 2.28  & \hspace {0.65cm} -4.43 & \hspace {0.65cm}  -1.74 & \hspace {0.65cm} -1.30\\
\hline
\end{tabular}
\caption{Difference between the {\it left-aligned} and the {\it right-aligned} constraint weights associated with each grammatical and information structure element. Weights are from the same three runs reported in Table \ref{table:weights}. Differences are computed as the {\it left-aligned} constraint weight minus the {\it right-aligned} constraint weight.}
\label{table:weightsdiffs}
\end{center}
\end{table}

%\begin{table}[h]
%\begin{center}
%\begin{tabular}{|l|p{2.5cm}|p{2.5cm}|p{2.5cm}|}
%\hline
%\multirow{2}{*}{Element} & \multicolumn{3}{|c|}{Sum of Left and Right Aligned Constraint Weights}\\
%& \hspace {.5cm} Run 1 & \hspace {.5cm} Run 2 & \hspace {.5cm} Run 3 \\
%\hline
%C-Topic & \hspace {.5cm} 22.81 & \hspace {.5cm}19.62 & \hspace {.5cm}21.64 \\
%Topic & \hspace {.5cm} 20.62 & \hspace {.5cm}17.19 & \hspace {.5cm}18.75 \\
%Focus & \hspace {.5cm} 18.75 & \hspace {.5cm}17.91 & \hspace {.5cm}17.55 \\
%Object & \hspace {.5cm} 15.62 & \hspace {.5cm}17.59 & \hspace {.5cm}15.90 \\
%Subject & \hspace {.5cm} 13.14 & \hspace {.5cm}16.25 & \hspace {.5cm}14.65 \\
%Verb & \hspace {.6cm}   9.08 & \hspace {.5cm}11.47 & \hspace {.5cm}11.50 \\
%\hline
%\end{tabular}
%\caption{Sum of the {\it left-aligned} and the {\it right-aligned} constraint weights associated with each grammatical and information structure element. Weights are from the same three runs reported in Table \ref{table:weights}}
%\label{table:weights2}
%\end{center}
%\end{table}

\subsection{Empirical Comparison with the GLA}\label{sec:glaresults}
As explained in Section \ref{sec:otandhg}, a Harmonic Grammar \citep{Legendre1990} uses the \emph{particular numeric values} of constraint weights to calculate scores, or harmonies, for input data and each possible word order, and the word order that scores the highest is chosen as the surface form. In contrast, the output from an Optimality Theory grammar is based only upon the \emph{relative, strict ranking} of the set of constraints. As explained in Section \ref{sec:learningcomp}, the Gradual Learning Algorithm learns numeric weights in a very similar way to the perceptron but the OT grammar that it learns makes word order predictions based upon a hierarchical OT ranking, which is inferred from the current weight vector. That is, the highest weighted constraint is the top ranked constraint, the second highest weighted is the second ranked constraint, and so on. In this Section, we empirically compare a model learned using the perceptron with a model learned using the GLA. The GLA and perceptron are very similar learning algorithms. Both make updates to weight vectors associated with constraints in the same way. However, they use different grammars (the GLA uses OT or SOT and the perceptron HG) to make predictions during training, which dictate which training examples are used to update the model. Because the learning algorithms are so similar, differences between prediction accuracies can be attributed to the difference between predicting word order using numeric (HG) constraint weights and using a hierarchical (OT) constraint ranking (see Section \ref{sec:gang} for examples and further discussion).

%In this section, we use the perceptron algorithm and the weight update function described in Section \ref{sec:learning} but make word order predictions based upon a strict, non-weighted ranking of the constraints, which is inferred from the current weight vector. That is, the highest weighted constraint is the top ranked constraint, the second highest weighted is the second ranked constraint, and so on. We refer to {\it OT predictions} as the word order predictions made by using the current weight vector to specify the order of strictly ranked constraints. 

We implemented the maximal GLA learning algorithm described by \cite{Boersma1997}. Like that implementation, we used a plasticity (learning rate) value of 0.01 and, using our development data, optimized the ranking spreading parameter, obtaining a value of 2.0.\footnote{In our experiments, we found that the plasticity values between .001 and .5 did not affect performance much on the development data one way or the other and performance decreased with higher and lower values. Similarly, with plasticity at .01, we found that relative spreading values between 1.0 and 4.0 resulted in the best performance, with lower and higher values negatively impacting KL-divergence, see below.} Plasticity is the magnitude of the weight added to or subtracted from constraint weights at each iteration of training. The relative spreading parameter is used in predicting surface forms. The product of the spreading parameter and some random noise\footnote{Gaussian distribution with mean 0, standard deviation 1.} is added to or subtracted from each constraint weight before the hierarchical OT relative rankings are inferred, which happens just before prediction. Like using the perceptron to learn a model, we iterated over the data several times before the GLA model converged.

In the formulation of Stochastic Optimality Theory given in \cite{Boersma1997}, noise is added to the current constraint ranking values (weights) before the learner predicts a surface form (in our case, a word order). Thus, in production, the learner chooses an expected word order from her current probability distribution over possible word orders rather than always predicting the most likely word order. Predicting the most likely word order, or using a maximum likelihood strategy, would minimize prediction errors, maximizing accuracy. In this section, we report results from using both the standard GLA strategy, henceforth \emph{SOT prediction}, and the maximum likelihood strategy, henceforth \emph{ML prediction} for both predicting test data surface forms and for predicting training data labels, which determines when updates to the ranking values are made. Table \ref{table:resultsGLA} shows the results.

%Repeated updates to the weight vector will result in changes in the corresponding strict constraint ranking, and the more violations of a particular type that are observed, the more likely the relevant constraint is to swap order with another constraint. As mentioned in Section \ref{sec:learningcomp}, this is in contrast to the constraint demotion algorithm, which swaps constraint orders as soon as evidence to do so is observed. If the strict ranking is changed with each mislabeled observation, the rankings are likely to oscillate back and forth because of the variation in our dataset. Making gradual weight updates will result in relatively stable strict constraint rankings and, additionally, will allow for tied constraints, the possibility of which the CD algorithm ignores. 

Tables \ref{table:results} and \ref{table:resultsGLA} show that even the best performing model learned by the GLA, which uses the ML strategy for prediction, does not perform nearly as well as the HG model learned by the perceptron. As mentioned above, this difference should be attributed to the difference between predicting word order using numeric (HG) constraint weights and using a hierarchical (OT or SOT) constraint ranking. 

\begin{table}[t]
\begin{center}
\begin{tabular}{|l|c|c|}
\hline
Training Prediction & Testing Prediction & Accuracy\\
\hline	
\multicolumn{2}{|l|}{Baseline} & 50.3\% \\
\multicolumn{2}{|l|}{Upper Bound} & 69.4\% \\
\hline
%HG System Performance & 67.0\% & 78.3\%\\
ML prediction & ML prediction & 59.7\% \\
ML prediction & SOT prediction & 47.4\% \\
SOT prediction & ML prediction & 59.7\% \\
SOT prediction & SOT prediction & 16.9\% \\ 
\hline
\multicolumn{2}{|l|}{Perceptron} & 67.0\% \\
\hline
\end{tabular}
\caption{GLA accuracies on a held out set of 1,000 test sentences. Training time prediction strategies and testing time prediction strategies are varied.}
\label{table:resultsGLA}
\end{center}
\end{table}

Table \ref{table:glaweights} shows the ranking values learned by the GLA when the SOT prediction strategy is used for predicting labels during training. For each grammatical and information structure element, the relative order of the left and right alignment constraints (e.g. Focus Right and Focus Left) is the same as the relative order between the two constraints in the first run of the learned perceptron weights in Table \ref{table:weights}.
In comparison with Table \ref{table:weights}, the range of the GLA weight values is very small. 
However, the two are not easily comparable since the HG weights are cumulative and the GLA weights are used to infer a hierarchical ranking. It is interesting to note, however, the differences between the ranking values compared with the spreading value, which is normalized to 1.0 in the second columns of weights. That is, for example, it is likely that enough noise will be added to the \emph{Object Right}, \emph{C-Topic Right}, or \emph{Subject Right} constraints such that their relative order will change. In contrast, it is unlikely the \emph{Focus Right} and \emph{Focus Left} constraint values would change enough for their relative orders to swap. It should also be noted that in ten runs, the relative ranking of the GLA-learned constraints did not vary at all. The ranking values shown in Table \ref{table:glaweights} are from one randomly chosen run.%, and the ranking value variances between runs are minimal.

\begin{table}[t]
\begin{center}
\begin{tabular}{|l|c|c|}
\hline
\multirow{2}{*}{Constraint} & Ranking Value & Ranking Value \\
& Weights normalized &  Spreading Value Normalized \\
\hline
C-Topic Left & 8.82 & 51.88 \\
Focus Right & 8.63 & 50.76 \\
Object Right & 8.55 & 50.29 \\
C-Topic Right & 8.53 & 50.18 \\
Subject Right & 8.49 & 49.94 \\
Topic Left & 8.42 & 49.53 \\
Subject Left & 8.40 & 49.41\\
Verb Left & 8.35 & 49.12\\
Object Left & 8.26 & 48.59\\
Verb Right & 7.96 & 46.82\\
Topic Right & 7.85 & 46.18\\
Focus Left & 7.76 & 45.65\\
\hline
\end{tabular}
\caption{Ranking values learned by the GLA algorithm. First, the weights are normalized to sum to 100. Then, the weights are normalized to have a spreading value of 1.}
\label{table:glaweights}
\end{center}
\end{table}

\subsection{Ganging-Up Effects}\label{sec:gang}

As mentioned in Section \ref{sec:glaresults}, the differences between the GLA-learned OT word order prediction accuracy and the perceptron-learned HG word order prediction accuracy can be attributed to the fact that the GLA uses a hierarchical set of constraints (an OT or SOT grammar) and the perceptron uses numeric constraints (an HG grammar). Since numeric weights can always express a hierarchical grammar, 
%\footnote{Weights would just need to be different enough that lower weighted constraints could never overpower higher ranked constraints, even cumulatively. The power of two series is one example of a series of weights that have this property: 1, 2, 4, 8, 16, 32, 64, 128, etc.}
it is important to question whether or not the additional expressivity of an HG grammar allows us to account for actual, observed grammar phenomena. 

Indeed, the model learned by the perceptron does outperform the GLA ML model in terms of accuracy, and in this section we closely examine some observed so-called \emph{ganging-up effects} which cause this increase in performance. That is, we present a concrete example of when the inferred hierarchical prediction (GLA) doesn't make the right prediction but allowing lower ranked constraints to gang-up on higher ranked constraints (perceptron) does result in the correct prediction.

One clear illustration of ganging-up effects is in predicting the input pattern of a topicalized subject and object and focused verb (line 7 of Table \ref{table:bigdv}). Given the GLA rankings shown in Table \ref{table:glaweights} and the perceptron weights shown in Table \ref{table:weights}, the GLA makes the maximum likelihood prediction (not inserting noise into the ranking values before prediction) of an SOV word order. In contrast, the perceptron weights predict SVO. In fact, as shown in Table \ref{table:bigdv}, SVO is the most likely word order for this input pattern in the training data. The same is true for the test data, making SVO the better prediction. Note that the weights learned in all of our perceptron runs predict SVO for this input, and we will use the weights given in the first run of Table \ref{table:weights} for illustration.

First, we show the GLA ML prediction in the tableau in Table \ref{table:glagang}. For convenience, we leave out \emph{C-Topic Left}, the highest ranked constraint, and \emph{C-Topic Right}, the fourth highest constraint, because, since there is no contrastively topicalized element, all word orders vacuously satisfy those two constraints. The OT tableau shows that all word orders except SOV and OSV are eliminated by the high ranked \emph{Focus Right} constraint and both SOV and OSV have the same violations on the \emph{Object Right}, \emph{Subject Right}, and \emph{Topic Left} constraints. OSV violates the \emph{Subject Left} constraint, leaving SOV as the predicted surface form. Note that we do not display the lower-ranked constraints, which are inconsequential to this analysis.

\begin{table}[h!]
\begin{center}
\begin{tabular}{c|l|c|c|c|c|c|}
\cline{2-7}
&  & Focus Right & Obj. Right & Subj. Right & Topic Left & Subj. Left \\
\cline{2-7}
& SVO & x & & x & & \\
& OVS & x & x & & & x \\
& VSO & x & & x & x & x \\
$\rightarrow$ & SOV & & x & x & & \\
& VOS & x & x & & x & x \\
& OSV & & x & x & & x \\
\cline{2-7}
\end{tabular}
\caption{GLA prediction of the input pattern topicalized subject, topicalized object, and focused verb. The GLA learned model, described above, predicts an SOV surface word order. Note that, for convenience, we leave out the \emph{C-Topic Left} and \emph{C-Topic Right} constraints since there is no contrastively topicalized element in sentences with this input pattern and all word orders vacuously satisfy them. We also leave out the constraints ranked below \emph{Subject Left} because they do not affect the analysis.}
\label{table:glagang}
\end{center}
\end{table}

Next, we show the perceptron-learned model's prediction for this input pattern. For simplicity, we only show the cumulative weights for SOV, predicted by the GLA-learned grammar, and SVO, which is the most harmonic word order under the perceptron-learned constraint weights. Table \ref{table:percgang} shows how the perceptron-learned harmonic grammar scores each word order. The four constraints for which the possible word orders have differing violation patterns are highlighted in bold. Like the GLA OT predictor, the \emph{Focus Right} constraint is the highest weighted constraint to discriminate between SOV and SVO. However, the table also shows that both \emph{Topic Right} and \emph{Object Right} have relatively high weights and also discriminate between the two surface word orders. Therefore, even though SVO violates \emph{Focus Right} and SOV does not, its final score is higher than that of SOV because it does not violate \emph{Topic Right} and \emph{Object Right}, and those constraints gang-up on the higher weighted \emph{Focus Right}.

\begin{table}[t]
\begin{center}
\begin{tabular}{|l|c||c|c||c|c|}
\hline
Constraint & Weight & SOV & SVO & SOV & SVO\\
\hline
C-Topic Left & 15.60  & 0 & 0 & & \\
Topic Left & 11.26  & + & + & & \\
\bf{Focus Right} & \bf{10.39}  & \bf{+} & \bf{--} & + 10.39 & -- 10.39 \\
\bf{Topic Right} & \bf{9.36}  & \bf{--} & \bf{+} & -- 9.36 & + 9.36 \\
\bf{Object Right} & \bf{8.63} & \bf{--} & \bf{+} & -- 8.63 & + 8.63\\
Focus Left & 8.36  & -- & -- & & \\
C-Topic Right & 7.21  & 0 & 0 & & \\
Subject Right & 7.18  & -- & -- & & \\
Object Left & 6.99  & -- & -- & & \\
Subject Left & 5.96  & + & + & & \\
Verb Left & 5.68  & -- & -- & & \\
\bf{Verb Right} & \bf{3.40}  & \bf{+} & \bf{--} & + 3.40 & -- 3.40 \\
\hline
\multicolumn{4}{|c|}{Sum of Differing Elements} & -- 4.2 & \bf{+ 4.2} \\
\hline
\end{tabular}
\caption{Perceptron prediction of the input pattern topicalized subject, topicalized object, and focused verb. The perceptron model, described above, predicts an SVO surface word order. Even though \emph{Focus Right} is the highest weighted constraint with violation patterns which are not the same for SOV and SVO, \emph{Topic Right} and \emph{Object Right} gang-up on it and allow SVO to win.}
\label{table:percgang}
\end{center}
\end{table}

This example of ganging-up effects shows how the expressivity of a Harmonic Grammar, which is beyond that of the hierarchical constraints in an Optimality Theory grammar, can result in a model which is capable of predicting correct surface forms that an OT model cannot predict. There are several additional input patterns in our small dataset and set of constraints that illustrate the same thing. These examples account for all of the over 7\% surface form prediction accuracy discrepancy between the perceptron learner and the GLA learner.

\subsection{Empirical Comparison with a MaxEnt Model}

We know in advance that the Constraint Demotion (CD) algorithm is incapable of learning a grammar from data that includes variation like ours, and we have shown empirically that the perceptron learner outperforms the GLA since the models it uses allow for ganging-up effects. In this section, we briefly compare the perceptron with a Maximum Entropy (MaxEnt) model, which also uses numerical weights and allows for the same effects. 

Table \ref{table:allresults} shows the prediction accuracies of the GLA (maximum likelihood predictions) OT grammar and perceptron HG grammar, both reported above, in addition to the accuracy of a MaxEnt model. We used the Natural Language Toolkit \citep{BirdKleinLoper09} implementation of a MaxEnt model and a conjugate gradient based algorithm to learn the grammar. The relatively small difference between the perceptron and MaxEnt accuracies is statistically significant.\footnote{Using a one sample Student t-test to compare the MaxEnt model's accuracy with the distribution of perceptron accuracies, estimated by sampling the performance of twenty perceptron-learned models. The difference is significant to the .0001 level.} %However, the difference is small, and we conclude only that the perceptron is capable of learning HG grammars that have prediction accuracies {\it at least as high as} MaxEnt models. 

\begin{table}[t]
\begin{center}
\begin{tabular}{|l|c|}
\hline
& Accuracy \\ %& Top-2 Accuracy \\
\hline
Baseline & 50.3\% \\ % & 68.6\% \\
GLA - ML Prediction & 59.7\% \\
MaxEnt &  66.5\%\\
Perceptron & 67.0\% \\ %& 78.3\%\\
Upper Bound & 69.4\% \\ % & 91.5\%\\
\hline
\end{tabular}
\caption{Baseline, GLA, MaxEnt, perceptron, and upper bound prediction strategy accuracies on a held out set of 1,000 test sentences.}
\label{table:allresults}
\end{center}
\end{table}

Table \ref{table:maxentweights} shows the weights associated with each attribute that are learned by the MaxEnt learner. Although it is somewhat difficult to interpret these weights, which are exponentiated in the model, we can compare their relative ranking with that of the weights learned by the perceptron. In comparison with the first run reported in Table \ref{table:weights}, the relative order of the left and right alignment constraints is the same for every grammatical and information structure element except for the subject. 

\begin{table}[h!]
\begin{center}
\begin{tabular}{|l|c|}
\hline
Constraint & Ranking Value \\
\hline
C-Topic Left & 2.30 \\
Subject Left & 0.53 \\
C-Topic Right & 0.39 \\
Object Right & 0.30 \\
Subject Right & 0.27 \\
Focus Right & 0.26 \\
Topic Left & 0.24 \\
Focus Left & 0.14 \\
Object Left & -0.04 \\
Verb Left & -0.18 \\
Topic Right & -1.00 \\
Verb Right & -1.33 \\
\hline
\end{tabular}
\caption{Attribute weights learned by the MaxEnt model learner.}
\label{table:maxentweights}
\end{center}
\end{table}

\section{Predicting Variation}
The attribute weights that the perceptron algorithm learns are intended to predict the single best word order in the set of possible orders. However, in this section we would like to predict a distribution over word orders and compare that distribution with the one observed for each input pattern (see Table \ref{table:bigdv}).\footnote{For example, we would like the learner to predict both VSO and SOV with high probability and SVO with lower probability when the S, V, and O are all focused elements, the last line in Table \ref{table:bigdv}.} %Therefore, we move to another online algorithm which is somewhat similar to the perceptron but which is capable of providing such a distribution over output labels. 

In order to use the perceptron's learned weights to predict such probability distributions, we use {\it Noisy HG}, which was proposed by \cite{boersmaPater08}. That is, for each constraint, we take a noise sample from a zero-mean Gaussian\footnote{Similar to our tuning for the GLA algorithm, we used our development set to tune a single parameter, the Gaussian variance, to .001} and add it to the current weight value. For each input pattern, we repeatedly sample and predict, keeping up with how many times each label is predicted. The sample predictions define a probability distribution over output labels. For both the GLA and the perceptron, we take 1,000 samples of the predicted word order for each input pattern.

Inserting noise into the perceptron-learned weights and the GLA-learned ranking values allows us to directly compare how well each predicts variation in our observed, held-out test data. As explained in Section \ref{sec:learningcomp}, the MaxEnt model naturally produces distributions over outputs, or surface word orders.

\subsection{Results}\label{sec:varresults}

KL-divergence is a measure of the difference between probability distributions \citep{kldivergence}. In order to compare the GLA-learned SOT grammar, perceptron-learned HG grammar, and the conjugate gradient learned MaxEnt grammar in terms of predicting data variation, we compute the KL-divergence between the true, empirically observed distribution over output labels in our test set and the distribution predicted by a given grammar. We compute this distance metric for each of the 27 input patterns for which there is some observed data. We then take a weighted average over the 27 inputs and output distributions based on the number of observed sentences in the test data. Table \ref{table:kldiv} presents these aggregate results. The GLA and perceptron algorithms are trained on 50 iterations over the training data and with the model parameters optimized as described above. The KL-divergence is averaged over ten runs. Smaller KL-divergence values indicate better estimates of the probability distributions. The perceptron HG grammar's average KL-divergence is lower than that of the GLA SOT grammar. The MaxEnt grammar has the smallest KL-divergence with the true distribution, meaning that it most accurately models variation in the data. 

\begin{table}
\begin{center}
\begin{tabular}{|l|c|c|}
\hline
Algorithm & Weighted $D_{KL}$ \\ %& $D_{KL}$ Variance\\
\hline
GLA & 0.92 \\ % 0.073  \\
Perceptron & 0.80 \\ %& 0.003 \\ 
MaxEnt & 0.53 \\ %& 0 \\
\hline
\end{tabular}
\caption{Weighted, by number of sentences with a given input pattern in the test data, KL-divergence between the true distribution and the predicted distributions. KL-divergence is averaged over ten runs. KL-divergence is measured in bits and smaller values are better than larger values.}
\label{table:kldiv}
\end{center}
\end{table}

In order to give some illustration of the probability distributions over surface word orders predicted by each algorithm, we show them for a few representative input patterns. Tables \ref{table:fffvar}, \ref{table:tffvar}, and \ref{table:ffcvar} show the true distribution over word orders as well as the distributions predicted by each model for different input patterns. It is interesting to see that all three algorithms seem to struggle to predict the same input patterns and easily predict the distribution over word orders for others. When all elements are focused, Table \ref{table:fffvar}, none of the three models does a good job of estimating the word order distribution. In particular, all three grossly underestimate the frequency of the SOV order. In contrast, when the subject and verb are focused and the object is contrastively topicalized, Table \ref{table:ffcvar}, all three models correctly predict that OVS is the most frequent surface form. 

The systematic errors exemplified in Table \ref{table:fffvar} indicate that the models may lack access to some type of important knowledge. That is, the models may benefit from additional constraints (attributes). \cite{zikanova06} makes a similar observation and concludes that we need to expand our theories of the relationship between information structure and sentence word orders in Czech in order to fully explain the data in the PDT. One benefit of the learning algorithms discussed here is that they can easily incorporate additional constraints and empirically test their effectiveness.
%claims that there are patterns of inconsistencies found in the Czech treebank  %with respect to the basic hypothesis about topicalized elements being first in sentence, followed by focused elements. Kept this project simple, would be easy to add additional proposed constraints from such analyses, from annotations. 

\begin{table}
\begin{center}
\begin{tabular}{|l|c|c|c|c|}
\hline
& True & GLA & Perceptron & MaxEnt \\
\hline
SVO & 16\% & 44\% & 39\% & 43\% \\
OVS & 5\% & 11\% & 26\% & 19\% \\
VSO & 31\% & 25\% & 19\% & 16\% \\
SOV & 47\% & 0\% & 2\% & 4\% \\
VOS & 1\% & 20\% & 14\% & 16\% \\
OSV & 1\% & 0\% & 1\% & 2\% \\
\hline
$D_{KL}$ & - & 5.56 & 2.00 & 1.59\\
\hline
\end{tabular}
\caption{True distribution over surface word orders for a focused subject, verb, and object along with the distribution over predictions made by the GLA SOT grammar, perceptron HG, and MaxEnt grammar.}
\label{table:fffvar}
\end{center}
\end{table}

\begin{table}
\begin{center}
\begin{tabular}{|l|c|c|c|c|}
\hline
& True & GLA & Perceptron & MaxEnt \\
\hline
SVO & 67\% & 65\% & 64\% & 62\% \\
OVS & 0\% & 5\% & 16\% & 4\% \\
VSO & 25\% & 17\% & 11\% & 21\% \\
SOV & 3\% & 0\% & 2\% & 6\% \\
VOS & 5\% & 10\% & 6\% & 3\% \\
OSV & 0\% & 3\% & 1\% & 3\% \\
\hline
$D_{KL}$ & - & 0.36 & 0.34 & 0.14\\
\hline
\end{tabular}
\caption{True distribution over surface word orders for a topicalized subject and focused verb and object.}
\label{table:tffvar}
\end{center}
\end{table}

\begin{table}
\begin{center}
\begin{tabular}{|l|c|c|c|c|}
\hline
& True & GLA & Perceptron & MaxEnt \\
\hline
SVO & 0\% & 11\% & 8\% & 10\% \\
OVS & 96\% & 82\% & 79\% & 74\% \\
VSO & 0\% & 5\% & 3\% & 4\% \\
SOV & 0\% & 1\% & 1\% & 1\% \\
VOS & 1\% & 1\% & 4\% & 3\% \\
OSV & 3\% & 1\% & 5\% & 8\% \\
\hline
$D_{KL}$ & - & 0.26 & 0.23 & 0.30\\
\hline
\end{tabular}
\caption{True distribution over surface word orders for a focused subject and verb and contrastively topicalized object.}
\label{table:ffcvar}
\end{center}
\end{table}

\begin{table}
\begin{center}
\begin{tabular}{|l|c|c|}
\hline
Predictor & $D_{KL}$ & Accuracy \\
\hline
Uniform distribution over word order labels & 1.53 & 16.7\% \\
Predict most likely single word order for input & 5.54 & 69.4\% \\ 
GLA prediction & 0.92 & 59.7\% \\
MaxEnt & 0.53 & 66.5\% \\
Perceptron prediction & 0.80 & 67.0\% \\
\hline
\end{tabular}
\caption{KL-divergence between the true distribution over word orders in a held out test set of 1,000 sentences and five strategies for predicting a distribution over word orders: a uniform distribution over word order labels, always predicting the most frequent word order label for a given input pattern (note that this strategy gives the upper bound on accuracy), and using the GLA, MaxEnt, and perceptron algorithms to learn models and make predictions.}% $D_{KL}$ is a weighted average over the input patterns and their frequencies in the held out set. Lower $D_{KL}$ scores are better than higher scores.}
\label{table:aggKLAcc}
\end{center}
\end{table}

Table \ref{table:aggKLAcc} gives the accuracy and the KL-divergence of each of the models discussed in this work, as well as two baseline strategies. The first baseline strategy always predicts a uniform distribution over word order labels, and the second baseline strategy only predicts the single most likely word order for each input pattern, which gives an upper bound on accuracy. All three of the learners perform well in terms of accuracy and predicting variation over output patterns in comparison with these two baselines. The perceptron HG model outperforms the GLA SOT model in both measures, which, as discussed at length in Section \ref{sec:gang}, indicates that the ganging-up effects are an important part of modeling this syntax phenomenon. The performance of the MaxEnt and perceptron models is similar. The perceptron grammar outperforms the MaxEnt model slightly with respect to accuracy and the MaxEnt model outperforms the perceptron grammar by a fair amount in predicting variation. The overall performance of the grammar learned by the perceptron is solid. Because of its simplicity and because it is an online learner, we conclude that it is an effective and elegant way to model grammar learning.

%\clearpage

\section{Conclusion}
In this work, we used a real dataset of annotated Czech sentences to explore the learnability of Czech word order. In particular, we explored the learnability of Optimality Theory and Harmonic Grammar models. We used basic ideas about the nature of Czech word order and corresponding alignment-based constraints on grammatical and information structure elements. By examining our dataset closely, we were able to define baseline and upper bound models for predicting word order and compared the models learned by several algorithms with them and each other. Given sufficient annotations, any additional theories of the nature of Czech word order could be incorporated into all of the learning models that we explored in the form of additional constraints. We could not only re-evaluate the models in light of different constraints, but we could evaluate the constraints and, thus, the theories themselves in terms of their contribution to the performance of the learning models.

By comparing several learning algorithms, including CD, GLA, perceptron, and a MaxEnt learner, we brought together two threads of research on OT/HG learnability. The first is learnability work that aims to model speaker and learner production. That is, the work in this area attempts to automatically learn models that mimic observed speaker production patterns, including variation. The computational modeling in this area is dominated by MaxEnt models. The second line of research aims to model the particular algorithms that speakers employ to learn a grammar. Because humans learn from one example sentence at a time, previous research in this area has used online models such as CD and the GLA. In this work, we performed the first empirical analysis of the perceptron as an HG learner. We showed that not only does the perceptron perform well in terms of modeling speaker production, including variation, but it is an attractive way to model the actual learning process since it is online and very simple. The perceptron has the added benefit of being a well-established algorithm in many fields, and its properties are well studied and understood.

In comparing the perceptron to the GLA, in particular, we showed empirically that using a Harmonic Grammar and allowing for ganging-up-effects results in a model that is more accurate than a similar OT model in terms of predicting surface forms. We showed one clear example of this effect in comparing a GLA-learned OT model and a perceptron-learned HG model. Three relatively high ranked constraints discriminated between two hypotheses. The OT model predicted the word order that did not violate the highest ranked constraint. However, this word order did violate the other two constraints, which both had relatively large HG weights. The HG model correctly predicted the word order which violated the highest ranked constraint but not the other two. That is, we saw that the two lower ranked constraints were able to effectively gang-up on the higher ranked constraint.

Finally, in addition to proposing, testing, and advocating for the perceptron as a model of HG learnability and, along the way, empirically showing that the additional complexity of HG models over OT models is necessary, we have shown that it is possible to computationally explore the learnability of syntax, not just phonology. This work opens the door for syntacticians to participate in the learnability discussion.

\clearpage

\appendix

\section{Learning Example}\label{sec:learningex}
This section gives a concrete example of how the perceptron algorithm is used to learn a predictor of Czech word order from the input data and the constraints described above. %For this example, I assume that the learning rate, $\eta$, is set to 1.
\begin{itemize}
\item{Randomly initialize weights: }
\begin{itemize}
%\item\small{$w_{sub\_left}=0$, $w_{sub\_right}=0$, ... ,  $w_{focus\_left}=0$ , $w_{focus\_right}=0$  \\
\item{The weight vector has elements corresponding to each of the 12 constraints: \\
\begin{table}[h!]\begin{flushright}\footnotesize\begin{tabular}{|c|c|c|c|c|c|c|c|c|c|c|c|}\hline
S-L & S-R & V-L & V-R & O-L & O-R & T-L & T-R & C-L & C-R & F-L & F-R \\
9 & 3 & 2 & 4 & 8 & 1 & 5 & 6 & 4 & 1 & 3 & 8 \\
\hline\end{tabular}\end{flushright}\end{table}

Where S stands for subject, V for verb or predicate, O for object, T for topic, C for contrastive topic, F for focus, L for aligned left, and R for aligned right.
}

\end{itemize}
\item{Input item 1: \\ x = Subj-Contrastive Topic, Verb-Focus, Obj-Focus, or SC-VF-OF \\ y = SVO}
\begin{itemize}
\item{For each word order label, determine attribute values and calculate score under the current model parameters}
\begin{itemize}

\item{SVO: \\ Attribute Vector $f(x,y)=f($SC-VF-OF, SVO$)$: }
\item[]{\begin{table}[h!]\begin{flushright}\scriptsize\begin{tabular}{|l|c|c|c|c|c|c|c|c|c|c|c|c|}\hline
 & S-L & S-R & V-L & V-R & O-L & O-R & T-L & T-R & C-L & C-R & F-L & F-R \\
$f(x,y)_i$ & 1 & -1 & -1 & -1 & -1 & 1 & 0 & 0 & 1 & -1 & -1 & 1 \\
\hline
$w_i$ & 9 & 3 & 2 & 4 & 8 & 1 & 5 & 6 & 4 & 1 & 3 & 8 \\
\hline\end{tabular}\end{flushright}\end{table}  }
\item[] {$Score = f($SC-VF-OF, SVO$)*w = {\bf 1}$ }

\item{OVS: \\ Attribute Vector $f(x,y)=f($SC-VF-OF, OVS$)$: }
\item [] {\begin{table}[h!!]\begin{flushright}\scriptsize\begin{tabular}{|l|c|c|c|c|c|c|c|c|c|c|c|c|}\hline
 & S-L & S-R & V-L & V-R & O-L & O-R & T-L & T-R & C-L & C-R & F-L & F-R \\
$f(x,y)_i$ & -1 & 1 & -1 & -1 & 1 & -1 & 0 & 0 & -1 & 1 & 1 & -1 \\
\hline
$w_i$ & 9 & 3 & 2 & 4 & 8 & 1 & 5 & 6 & 4 & 1 & 3 & 8 \\
\hline\end{tabular}\end{flushright}\end{table}  }
\item[]{$Score = f($SC-VF-OF, OVS$)*w = {\bf -13}$ }

\item{VSO: \\ Attribute Vector $f(x,y)=f($SC-VF-OF, VSO$)$: }
\item [] {\begin{table}[h!!]\begin{flushright}\scriptsize\begin{tabular}{|l|c|c|c|c|c|c|c|c|c|c|c|c|}\hline
 & S-L & S-R & V-L & V-R & O-L & O-R & T-L & T-R & C-L & C-R & F-L & F-R \\
$f(x,y)_i$ & -1 & -1 & 1 & -1 & -1 & 1 & 0 & 0 & -1 & -1 & 1 & 1 \\
\hline
$w_i$ & 9 & 3 & 2 & 4 & 8 & 1 & 5 & 6 & 4 & 1 & 3 & 8 \\
\hline\end{tabular}\end{flushright}\end{table}  }
\item[]{$Score = f($SC-VF-OF, VSO$)*w = {\bf -15}$ }

\item{SOV: \\ Attribute Vector $f(x,y)=f($SC-VF-OF, SOV$)$: }
\item [] {\begin{table}[h!!]\begin{flushright}\scriptsize\begin{tabular}{|l|c|c|c|c|c|c|c|c|c|c|c|c|}\hline
 & S-L & S-R & V-L & V-R & O-L & O-R & T-L & T-R & C-L & C-R & F-L & F-R \\
$f(x,y)_i$ & 1 & -1 & -1 & 1 & -1 & -1 & 0 & 0 & 1 & -1 & -1 & 1 \\
\hline
$w_i$ & 9 & 3 & 2 & 4 & 8 & 1 & 5 & 6 & 4 & 1 & 3 & 8 \\
\hline\end{tabular}\end{flushright}\end{table}  }
\item[]{$Score = f($SC-VF-OF, SOV$)*w = {\bf 7}$ }

\clearpage

\item{VOS: \\ Attribute Vector $f(x,y)=f($SC-VF-OF, VOS$)$: }
\item [] {\begin{table}[h!!]\begin{flushright}\scriptsize\begin{tabular}{|l|c|c|c|c|c|c|c|c|c|c|c|c|}\hline
 & S-L & S-R & V-L & V-R & O-L & O-R & T-L & T-R & C-L & C-R & F-L & F-R \\
$f(x,y)_i$ & -1 & 1 & 1 & -1 & -1 & -1 & 0 & 0 & -1 & 1 & 1 & -1 \\
\hline
$w_i$ & 9 & 3 & 2 & 4 & 8 & 1 & 5 & 6 & 4 & 1 & 3 & 8 \\
\hline\end{tabular}\end{flushright}\end{table}  }
\item[]{$Score = f($SC-VF-OF, VOS$)*w = {\bf -25}$ }

\item{OSV: \\ Attribute Vector $f(x,y)=f($SC-VF-OF, OSV$)$: }
\item [] {\begin{table}[h!!]\begin{flushright}\scriptsize\begin{tabular}{|l|c|c|c|c|c|c|c|c|c|c|c|c|}\hline
 & S-L & S-R & V-L & V-R & O-L & O-R & T-L & T-R & C-L & C-R & F-L & F-R \\
$f(x,y)_i$ & -1 & -1 & -1 & 1 & 1 & -1 & 0 & 0 & -1 & -1 & 1 & 1 \\
\hline
$w_i$ & 9 & 3 & 2 & 4 & 8 & 1 & 5 & 6 & 4 & 1 & 3 & 8 \\
\hline\end{tabular}\end{flushright}\end{table}  }
\item[]{$Score = f($SC-VF-OF, OSV$)*w = {\bf 3}$ }

\end{itemize}

\item{Predict the label with the highest score: SOV}
\item{$SVO \neq SOV$ \\ Update the weights on attributes with differing values: \\ $f($SC-VF-OF, SVO$) - f($SC-VF-OF, SOV$)$}
\begin{itemize}
\item [] {\begin{table}[h!!]\begin{flushright}\scriptsize\begin{tabular}{|l|c|c|c|c|c|c|c|c|c|c|c|c|}\hline
 & S-L & S-R & V-L & V-R & O-L & O-R & T-L & T-R & C-L & C-R & F-L & F-R \\
SVO & 1 & -1 & -1 & -1 & -1 & 1 & 0 & 0 & 1 & -1 & -1 & 1 \\
SOV & 1 & -1 & -1 & 1 & -1 & -1 & 0 & 0 & 1 & -1 & -1 & 1 \\
SVO-SOV & 0 & 0 & 0 & - & 0 & + & 0 & 0 & 0 & 0 & 0 & 0 \\
\hline
$w_t$ & 9 & 3 & 2 & 4 & 8 & 1 & 5 & 6 & 4 & 1 & 3 & 8 \\
\hline
$w_{t+1}$ & 9 & 3 & 2 & {\bf 3} & 8 & {\bf 2} & 5 & 6 & 4 & 1 & 3 & 8 \\
\hline\end{tabular}\end{flushright}\end{table}  }

\end{itemize}
\end{itemize}

\end{itemize}

\clearpage

\bibliographystyle{spbasic}     
\bibliography{Irvine_Czech_OT}

\begin{thebibliography}{53}
\providecommand{\natexlab}[1]{#1}
\providecommand{\url}[1]{{#1}}
\providecommand{\urlprefix}{URL }
\expandafter\ifx\csname urlstyle\endcsname\relax
  \providecommand{\doi}[1]{DOI~\discretionary{}{}{}#1}\else
  \providecommand{\doi}{DOI~\discretionary{}{}{}\begingroup
  \urlstyle{rm}\Url}\fi
\providecommand{\eprint}[2][]{\url{#2}}

\bibitem[{Bird et~al(2009)Bird, Klein, and Loper}]{BirdKleinLoper09}
Bird S, Klein E, Loper E (2009) Natural Language Processing with Python:
  Analyzing Text with the Natural Language Toolkit. O'Reilly, Beijing

\bibitem[{Boersma(1997)}]{Boersma1997}
Boersma P (1997) How we learn variation, optionality, and probability. In:
  Proceedings of the Institute of Phonetic Studies, vol~21, pp 43--58

\bibitem[{Boersma and Hayes(2001)}]{Boersma2001}
Boersma P, Hayes B (2001) Empirical tests of the gradual learning algorithm.
  Linguistic Inquiry 32(1):45--86

\bibitem[{Boersma and Pater(2008)}]{boersmaPater08}
Boersma P, Pater J (2008) Convergence properties of a gradual learning
  algorithm for harmonic grammar. Rutgers Optimality Archive 970

\bibitem[{Bresnan(2007)}]{Bresnan07}
Bresnan J (2007) Is syntactic knowledge probabilistic? {E}xperiments with the
  english dative alternation. Roots: Linguistics in Search of Its Evidential
  Base pp 77--96

\bibitem[{Bur\'{a}\v{n}ov\'{a} et~al(2000)Bur\'{a}\v{n}ov\'{a},
  Haji\v{c}ov\'{a}, and Sgall}]{buranova00}
Bur\'{a}\v{n}ov\'{a} E, Haji\v{c}ov\'{a} E, Sgall P (2000) Tagging of very
  large corpora: topic-focus articulation. In: Proceedings of the 18th
  Conference on Computational Linguistics, pp 139--144

\bibitem[{Choi(1999)}]{Choi99}
Choi HW (1999) {Optimizing structure in context: Scrambling and information
  structure}. CSLI Publ.

\bibitem[{Coetzee and Pater(2008)}]{coetzee08a}
Coetzee A, Pater J (2008) Weighted constraints and gradient restrictions on
  place co-occurrence in {M}una and {A}rabic. Natural Language {\&} Linguistic
  Theory 26:289--337

\bibitem[{Coetzee and Pater(to appear)}]{coetzee08b}
Coetzee A, Pater J (to appear) The place of variation in phonological theory.
  In: Goldsmith J, Riggle J, Yu A (eds) The Handbook of Phonological Theory,
  2nd edn, Oxford: Blackwell

\bibitem[{Collins(2002)}]{Collins:2002}
Collins M (2002) Discriminative training methods for hidden markov models:
  theory and experiments with perceptron algorithms. In: Proceedings of the
  Conference on Empirical Methods in Natural Language Processing - Volume 10,
  Association for Computational Linguistics, Stroudsburg, PA, USA, pp 1--8

\bibitem[{Costa(1997)}]{Costa97}
Costa J (1997) Word order and constraint interaction. Seminarios de Linguistica

\bibitem[{Costa(2001)}]{Costa}
Costa J (2001) The emergence of unmarked word order. In: Legendre G, Grimshaw
  J, Vikner S (eds) Optimality-Theoretic Syntax, MIT Press

\bibitem[{Doyle and Levy(2008)}]{Doyle08}
Doyle G, Levy R (2008) Environment prototypicality in syntactic alternation.
  In: Proceedings of the 34th Annual Meeting of the Berkeley Linguistics
  Society

\bibitem[{Eisner(2000)}]{eisner00}
Eisner J (2000) Easy and hard constraint ranking in {O}ptimality {T}heory:
  Algorithms and complexity. In: Eisner J, Karttunen L, Th\'{e}riault A (eds)
  Finite-State Phonology: Proceedings of the 5th Workshop of the {ACL}
  {S}pecial {I}nterest {G}roup in Computational Phonology ({SIGPHON}),
  Luxembourg, pp 22--33

\bibitem[{Goldwater and Johnson(2003)}]{Goldwater2003}
Goldwater S, Johnson M (2003) Learning {OT} constraint rankings using a maximum
  entropy model. In: Proceedings of the Stockholm Workshop on Variation within
  Optimality Theory, pp 111 -- 120

\bibitem[{Hajic et~al(2001)Hajic, Hladka, and Pajas}]{pdt}
Hajic J, Hladka B, Pajas P (2001) The {P}rague {D}ependency {T}reebank:
  Annotation structure and support. In: In Proc. of the IRCS Workshop on
  Linguistic Databases, pp 105--114

\bibitem[{Haji{\v{c}}ov{\'a} et~al(1995)Haji{\v{c}}ov{\'a}, Sgall, and
  Skoumalova}]{Hajicova}
Haji{\v{c}}ov{\'a} E, Sgall P, Skoumalova H (1995) An automatic procedure for
  topic-focus identification. Computational Linguistics 21(1):81--94

\bibitem[{Haji{\v{c}}ov{\'a} et~al(1998)Haji{\v{c}}ov{\'a}, Partee, and
  Sgall}]{Hajicova1998}
Haji{\v{c}}ov{\'a} E, Partee BH, Sgall P (1998) {Topic-focus articulation,
  tripartite structures, and semantic content}. Amsterdam:Kluwer

\bibitem[{Hayes and Wilson(2008)}]{HayesWilson08}
Hayes B, Wilson C (2008) A maximum entropy model of phonotactics and
  phonotactic learning. Linguistic Inquiry pp 379--440

\bibitem[{Jaeger and Rosenbach(2003)}]{Jaeger2003}
Jaeger G, Rosenbach A (2003) The winner takes it all {--} almost.
  {C}umulativity in grammatical variation. In: Proceedings of the Workshop on
  Optimality Theoretic Syntax

\bibitem[{Jarosz(2006)}]{Gaja}
Jarosz G (2006) Richness of the base and probabilistic unsupervised learning in
  optimality theory. In: Proceedings of the Eighth Meeting of the ACL Special
  Interest Group on Computational Phonology at HLT-NAACL

\bibitem[{Jesney and Tessier(2009)}]{jesney2009}
Jesney K, Tessier AM (2009) Gradual learning and faithfulness: consequences of
  ranked vs. weighted constraints. Proceedings of the 38th Meeting of the North
  East Linguistics Society (NELS 38) 1:375--388

\bibitem[{Johnson(2009)}]{Johnson2009}
Johnson M (2009) How the statistical revolution changes (computational)
  linguistics. In: Proceedings of the European Association for Computational
  Linguistics Workshop on the Interaction between Linguistics and Computational
  Linguistics

\bibitem[{Khardon et~al(2005)Khardon, Wachman, and Collins}]{Khardon05}
Khardon R, Wachman G, Collins J (2005) Noise tolerant variants of the
  {P}erceptron algorithm. Journal of Machine Learning Research 8:227--248

\bibitem[{King(1995)}]{king}
King TH (1995) Configuring topic and focus in {R}ussian. In: CSLI Dissertations
  in Linguistics, Cambridge University Press

\bibitem[{Kiss(1995)}]{kiss}
Kiss K (1995) Discourse configurational languages. Oxford studies in
  comparative syntax, Oxford University Press

\bibitem[{Kullback and Leibler(1951)}]{kldivergence}
Kullback S, Leibler RA (1951) On information and sufficiency. The Annals of
  Mathematical Statistics 22(1):79--86

\bibitem[{Lambrecht(1994)}]{lambrecht}
Lambrecht K (1994) Information Structure and Sentence Form: Topic, Focus, and
  the Mental Representations of Discourse Referents. Cambridge University Press

\bibitem[{Legendre et~al(1990{\natexlab{a}})Legendre, Miyata, and
  Smolensky}]{Legendre1990}
Legendre G, Miyata Y, Smolensky P (1990{\natexlab{a}}) Can connectionism
  contribute to syntax? {H}armonic {G}rammar, with an application. In:
  Proceedings of the 26th Meeting of the Chicago Linguistic Society

\bibitem[{Legendre et~al(1990{\natexlab{b}})Legendre, Miyata, and
  Smolensky}]{legendre1990a}
Legendre G, Miyata Y, Smolensky P (1990{\natexlab{b}}) Harmonic grammar Ñ a
  formal multi level connectionist theory of linguistic well formedness: An
  application. In: Proceedings of the Twelfth Annual Conference of the
  Cognitive Science Society, Cambridge, MA, p 884Ð891

\bibitem[{Legendre et~al(1990{\natexlab{c}})Legendre, Miyata, and
  Smolensky}]{legendre1990b}
Legendre G, Miyata Y, Smolensky P (1990{\natexlab{c}}) Harmonic grammar Ñ a
  formal multi level connectionist theory of linguistic well formedness:
  Theoretical foundations. In: Proceedings of the Twelfth Annual Conference of
  the Cognitive Science Society, Cambridge, MA, p 388Ð395

\bibitem[{Legendre et~al(2004)Legendre, Hagstrom, Chen-Main, Tao, and
  Smolensky}]{Legendre04}
Legendre G, Hagstrom P, Chen-Main J, Tao L, Smolensky P (2004) Deriving output
  probabilities in child {M}andarin from a dual-optimization grammar. Lingua
  114:1147--1185

\bibitem[{Legendre et~al(2006{\natexlab{a}})Legendre, Miyata, and
  Smolensky}]{harmonicmindchaptera}
Legendre G, Miyata Y, Smolensky P (2006{\natexlab{a}}) The interaction of
  syntax and semantics: A harmonic grammar account of split intransitivity. In:
  Smolensky P, Legendre G (eds) The Harmonic Mind: From Neural Computation to
  Optimality-Theoretic Grammar Volume 1, The MIT Press, pp 417--452

\bibitem[{Legendre et~al(2006{\natexlab{b}})Legendre, Sorace, and
  Smolensky}]{harmonicmindchapterb}
Legendre G, Sorace A, Smolensky P (2006{\natexlab{b}}) The optimality
  theory-harmonic grammar connection. In: Smolensky P, Legendre G (eds) The
  Harmonic Mind: From Neural Computation to Optimality-Theoretic Grammar Volume
  2, The MIT Press, pp 339--402

\bibitem[{Littlestone(1988)}]{Littlestone:1988}
Littlestone N (1988) Learning quickly when irrelevant attributes abound: A new
  linear-threshold algorithm. Machine Learning 2:285--318

\bibitem[{Magri(2010)}]{Giorgio10}
Magri G (2010) {HG} has no computational advantages over {OT}: consequences for
  the theory of {OT} online algorithms. Manuscript, IJN, ENS Available as
  ROA-1104

\bibitem[{Magri(2011)}]{magri2011}
Magri G (2011) {HG} has no computational advantages over {OT}: new tools for
  computational {OT}. Linguistic Inquiry

\bibitem[{McCarthy(2003)}]{mccarthy03}
McCarthy JJ (2003) {OT} constraints are categorical. Phonology 20:75--138

\bibitem[{Naughton(2012)}]{naughton2012czech}
Naughton J (2012) Czech: An Essential Grammar. Routledge essential grammars,
  Taylor \& Francis,
  \urlprefix\url{http://books.google.com/books?id=jDvpsLRE0pkC}

\bibitem[{Pater(2008)}]{Pater08}
Pater J (2008) Gradual learning and convergence. Linguistic Inquiry
  39(2):334--345

\bibitem[{Pater(2009)}]{pater2009}
Pater J (2009) Weighted constraints in generative linguistics. Cognitive
  Science 33(6):999--1035

\bibitem[{Pater et~al(2007)Pater, Bhatt, and Potts}]{PaterBhattPotts07}
Pater J, Bhatt R, Potts C (2007) Linguistic optimization, ms., UMass Amherst.
  ROA 924-0907

\bibitem[{Pater et~al(2010)Pater, Smith, Staubs, Jesney, and Mettu}]{pater10}
Pater J, Smith D, Staubs R, Jesney K, Mettu R (2010) Learning hidden structure
  with a log-linear model of grammar probabilities. In: Proceedings of the
  Annual Meeting of the Linguistic Society of America

\bibitem[{Potts et~al(2010)Potts, Pater, Jesney, Bhatt, and Becker}]{potts2010}
Potts C, Pater J, Jesney K, Bhatt R, Becker M (2010) Harmonic grammar with
  linear programming: From linear systems to linguistic typology. Phonology
  27(1):1--41

\bibitem[{Prince and Smolensky(1993)}]{Prince1993}
Prince A, Smolensky P (1993) Optimality theory: Constraint interaction in
  generative grammar. In: Rutgers University Center for Cognitive Science
  Technical Report

\bibitem[{Prince and Smolensky(2004)}]{prince2004}
Prince A, Smolensky P (2004) Optimality Theory: Constraint Interaction in
  Generative Grammar. Blackwell Publishers

\bibitem[{Rosenblatt(1958)}]{rosenblatt1958}
Rosenblatt F (1958) The perceptron: a probabilistic model for information
  storage and organization in the brain. Psychological Reviews 65(6):386--408

\bibitem[{Tesar(1995)}]{tesarthesis}
Tesar B (1995) Computational optimality theory. Doctoral dissertation,
  Department of Computer Science, University of Colorado at Boulder, Boulder,
  CO, rutgers Optimality Archive 90

\bibitem[{Tesar and Smolensky(1993)}]{tesarsmolensky1993}
Tesar B, Smolensky P (1993) The learnability of optimality theory: An algorithm
  and some basic complexity results. Tech. Rep. CU-CS-678-93, University of
  Colorado at Boulder, Boulder, CO, rutgers Optimality Archive 2

\bibitem[{Tesar and Smolensky(1998)}]{Tesar1998}
Tesar B, Smolensky P (1998) Learnability in optimality theory. Linguistic
  Inquiry 29(2):229--268

\bibitem[{Tesar and Smolensky(2000)}]{Tesar2000}
Tesar B, Smolensky P (2000) Learnability in Optimality Theory. MIT Press

\bibitem[{Wilson(2006)}]{Wilson06}
Wilson C (2006) Learning phonology with substantive bias: An experimental and
  computational study of velar palatalization. Cognitive Science 30:945--982

\bibitem[{Zik\'{a}nov\'{a}(2006)}]{zikanova06}
Zik\'{a}nov\'{a} {\v{S}} (2006) What do the data in {PDT} say about systemic
  ordering in {C}zech? The Prague Bulletin of Mathematical Linguistics
  86:39--46

\end{thebibliography}

\end{document}